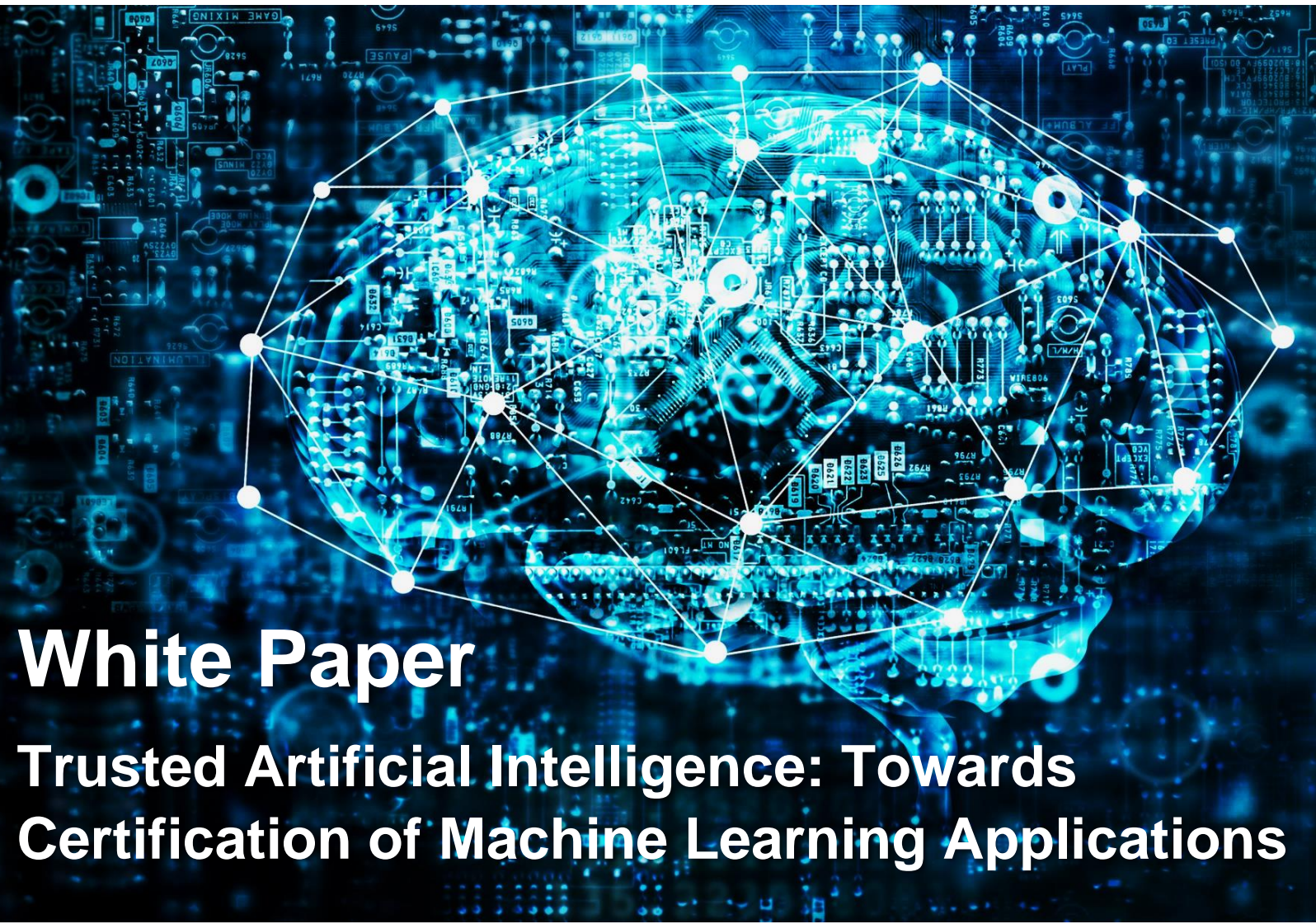

# White Paper

## Trusted Artificial Intelligence: Towards Certification of Machine Learning Applications

# Trusted Artificial Intelligence: Towards Certification of Machine Learning Applications

*Vienna, March 17th, 2021*

*Philip Matthias Winter* [2]
*Sebastian Eder* [1]
*Johannes Weissenböck* [1]
*Christoph Schwald* [1]
*Thomas Doms* [1]
*Tom Vogt* [1]
*Sepp Hochreiter* [2]
*Bernhard Nessler* [2]

[1] **TÜV AUSTRIA Group**, [2] **Johannes Kepler University Linz – Institute for Machine Learning**

# Reviewer statements

*"The white paper presents initial approaches to the certification of ML systems. An essential part of the paper is a catalog for auditing an ML system, which has not yet been presented in this form. It is particularly positive that existing criteria from software development are integrated into the audit catalog and that ethical issues are also addressed. The catalog is still limited to certain problem areas such as supervised learning procedures and applications of low criticality, but lays the foundation for more extensive certification efforts."*

**Prof. Dr.-Ing. Marco Huber**

Cyber Cognitive Intelligence (CCI)
Fraunhofer Institute for Manufacturing Engineering and Automation (IPA)

*"Data-driven artificial intelligence such as Deep Learning poses new challenges for technical standards specification and certification. Given still ongoing research and some unresolved issues, this article helps raise awareness of this issue for practitioners and serves as a useful starting point for a more in-depth study of this topic. In this way, key concepts are made comprehensible and differences to conventional engineering are highlighted. Very helpful in this context are the guidelines presented in the guise of an audit catalog for avoiding pitfalls in daily work with machine and deep learning."*

**Priv.-Doz. Dr. Bernhard A. Moser**

President of Austrian Society of Artificial Intelligence (ASAI)
and Research Director of Software Competence Center Hagenberg (SCCH)

# Abstract

Artificial Intelligence is one of the fastest growing technologies of the 21st century and accompanies us in our daily lives when interacting with technical applications. However, reliance on such technical systems is crucial for their widespread applicability and acceptance. The societal tools to express reliance are usually formalized by lawful regulations, i.e., standards, norms, accreditations, and certificates. Therefore, the TÜV AUSTRIA Group in cooperation with the Institute for Machine Learning at the Johannes Kepler University Linz, proposes a certification process and an audit catalog for Machine Learning applications. We are convinced that our approach can serve as the foundation for the certification of applications that use Machine Learning and Deep Learning, the techniques that drive the current revolution in Artificial Intelligence. While certain high-risk areas, such as fully autonomous robots in workspaces shared with humans, are still some time away from certification, we aim to cover low-risk applications with our certification procedure. Our holistic approach attempts to analyze Machine Learning applications from multiple perspectives to evaluate and verify the aspects of secure software development, functional requirements, data quality, data protection, and ethics. Inspired by existing work, we introduce four criticality levels to map the criticality of a Machine Learning application regarding the impact of its decisions on people, environment, and organizations. Currently, the audit catalog can be applied to low-risk applications within the scope of supervised learning as commonly encountered in industry. Guided by field experience, scientific developments, and market demands, the audit catalog will be extended and modified accordingly.

# *Content*

# 1 Introduction

Currently, Artificial Intelligence (AI) is addressed in the roadmap of almost every company, even if its business has no technological focus. Current success stories of AI create huge expectations, which are further reinforced by advances in digitalization and hardware breakthroughs. Machine Learning (ML) and more specifically Deep Learning (DL) has emerged as a core technology of AI and is driving the AI revolution. Consequently, the number of products using ML is rising steadily.

However, alongside with the promises of AI, its corresponding risks must be considered and mitigated appropriately. Utilizing ML technologies on a large scale will only be possible if humans that are affected by these technologies have reliance on their safety and security. For a new technology to be successful, reliance on it is key [1, 2]. We rather use the term *reliance* than the term *trust*, which is often used in public discussions since the term *reliability* is more related to technical products and, therefore, is more appropriate for certification [3]. Moreover, the term *trust* has different meanings in fields less related to AI, e.g. cryptography.

Certification is used to confirm whether certain requirements (such as technical standards), e.g., for products, persons, or processes have been met. During the certification process, an independent third party confirms compliance with these requirements. The evaluation comprises both qualitative and quantitative requirements. Note that the trust in the competence of the auditors and their experienced judgments play a crucial role in most certification processes. If the process is concluded with a positive result, conformity [4] is given, i.e., the requirements are fulfilled. Non-conformity thus implies the inadequate fulfillment of certain, precisely defined requirements.

In public, ML systems are often falsely perceived as black boxes whose decisions are not comprehensible. Although Deep Learning models are certainly complex, they are not black boxes. In fact, it would be more accurate to refer to them as glass boxes, because we can literally look inside and see what each component is doing (quote from [5]). Every decision and every numerical operation of an ML system can in principle be reproduced and analyzed, yet the complex interactions between these operations makes it infeasible for humans to comprehend the decision process as a whole.

At the same time, ML is more accessible than ever before. Many companies advertise their own "AutoML" tools in which ML systems can be readily constructed with user-friendly drag-and-drop actions. While this offers certain advantages, it is essential to know the proper scientific methodologies. Often errors emerge very easily when applying ML techniques, e.g., due to lack of experience that leads to incorrect usage of ML methods. These errors are often difficult to detect afterwards. For example, one may misinterpret the estimate of generalization performance of an ML model and report the training set performance instead of the test set performance. If not captured before deployment, such errors can have far-reaching and hazardous consequences. Therefore, an a-priori distrust towards such new systems might be justified to a certain extent, not due to the principles of the applied methods but due to the possible lack of competence of the application developer that uses them. Even though there are commonly known best practices, the field still lacks well-established standards and guidelines.

Sophisticated mathematical proofs and theorems of ML algorithm properties, together with a well-defined statistical analysis of the specific implementation, are the tools for guarantees in the world of ML. Building up both competence and trust that results from high reliance first and foremost requires communicating at least the relevant conclusions of the proofs and statistical principles in comprehensible ways. Societal tools to express this trust are formalized by lawful regulations, accreditations, and certificates, which will subsequently serve as guidelines for developers in order to build reliable applications.

The overall goal of this paper is threefold: Firstly, we aim to clarify important ML principles in simple terms in order to reach a broad audience. Secondly, we aim to discuss important ML-related aspects and challenges which are relevant in the context of certification. Thirdly, we aim to utilize existing certification procedures to take a first step towards developing a certification procedure for ML applications.

In this paper, we introduce a certification approach, where an independent third party such as TÜV AUSTRIA testifies the quality and safe usage of ML applications. Certifications are quality seals for products and services and often prerequisite for admission to the market. Certifications serve as guidelines for developers in order to build reliable applications and create confidence for buyers, decision-makers, and clients. Consequently, this work mainly addresses developers, solution providers, and buyers of technical services in a Business-to-Business and a Business-to-Consumer context. Certification of ML applications raises many challenging problems in practice [6, 7]. Since well-established certification procedures for classical software cannot be adopted in a straight-forward manner, one has to reconsider major topics of certification in the context of ML. We believe that a broad certification approach of ML applications will play a significant role to raise the overall quality and safe usage of this technology and thus increase public acceptance and reliance.

Currently, our certification approach is restricted to supervised learning tasks with a low-risk potential. We focus on supervised learning since it is already heavily used, and thus highly important, for many technological applications in industry. Moreover, supervised learning can be expressed by the means of well-defined problem definitions. Using these problem definitions, it is possible to come up with well-defined rules that evaluate and rate specific methods, algorithms and realizations that emerge from this framework. These provably correct mathematical rules allow for developing a certification process that ensures a certain level of quality and reliability for the algorithm under test. Our focus on certification of ML applications with low-risk potential is mainly motivated by the urgent industrial need in this field.

In this paper, we also introduce an audit catalog which serves as a foundation for certifying ML applications. With field experience (from applying the audit catalog in practice) as well as utilizing insights from scientific developments (such as aspects related to data integrity and stability of ML algorithms [8, 9, 10]), the audit catalog will be extended, modified, refined, and formalized accordingly.

As we have now motivated the urgent need for the certification of ML applications, we would like to outline the structure of this work. In Section 2, we provide a brief overview and describe important concepts in the field of AI. In Section 3, we outline ongoing activities regarding standardization and certification of ML applications. We also discuss the topics reproducibility, interpretability, and key challenges to motivate a reliable quality assessment approach for ML applications. In Section 4, we present our certification approach and our audit catalog for ML applications. We demonstrate the usage of the catalog via typical scenarios, revealing some common pitfalls and mistakes in ML. Lastly, we conclude our findings and provide a brief outlook on possible future directions in Section 5.

# 2   Background of Artificial Intelligence

In this section we give a short overview of important concepts in AI and address current research developments. For a brief summary of the history of AI, we refer the reader to the Appendix.

## 2.1 Definitions and Learning

In the following, we describe the most important concepts of AI.

**Artificial Intelligence (AI):** The term *Artificial Intelligence* was coined by John McCarthy in the "Dartmouth Summer Project on Artificial Intelligence" in 1956 [11]. A definition of AI was given by Barr & Feigenbaum [12] in 1981:

> "Artificial Intelligence (AI) is the part of computer science concerned with designing intelligent computer systems, that is, systems that exhibit characteristics we associate with intelligence in human behavior – understanding language, learning, reasoning, solving problems, and so on."

Currently, AI refers to the ability of machines to perform cognitive tasks commonly associated with human intelligence including perception, learning, reasoning, planning, speech and language, and taking actions. In particular, AI also includes tasks of robotics, i.e., the ability to move and manipulate objects and autonomous orientation and motion.

**Machine Learning (ML):** The term *Machine Learning* was introduced by Arthur Samuel [13] in 1959. In contrast to deductive reasoning (like writing a program), ML relies on inductive reasoning (learning from data). The overall goal is to learn general principles from observed data. A definition of ML was given by Tom Mitchel [14] in 1997:

> "A computer program is said to learn from experience E with respect to some class of tasks T, and performance measure P, if its performance at tasks in T, as measured by P, improves with experience E."

Historically, ML was divided into three main areas: supervised learning, unsupervised learning, and reinforcement learning. Nowadays these areas are no longer so clearly distinguishable from each other and many more sub-areas such as weakly-supervised learning, semi-supervised learning, self-supervised learning, or active learning have been established. Moreover, for practical applications, often ML techniques from different sub-areas are combined. Figure 1 shows the different sub-areas of ML, which are discussed in more detail below.

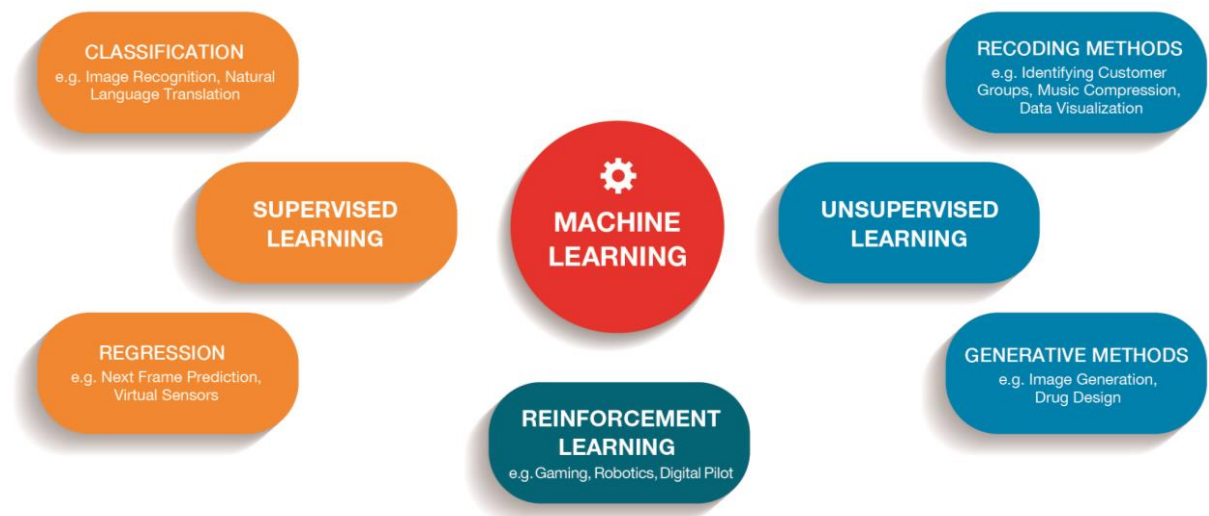

Figure 1: Examples of different subareas of ML.

**Deep Learning (DL):** The term *Deep Learning* was introduced in the work of Aizenberg et al. [15] in 2000. DL is a subcategory of ML that makes use of ANNs. More general definition was given by LeCun et al. [16] in 2015:

> "Deep-learning methods are representation-learning methods with multiple levels of representation, obtained by composing simple but non-linear modules that each transform the representation at one level (starting with the raw input) into a representation at a higher, slightly more abstract level."

Multi-layered artificial neural networks (ANNs) are the typical framework for the implementation of deep-learning models (an ANN and how it relates to the DL definition is explained in the example below). ANNs can be categorized into feed-forward networks [17], convolutional networks [18, 19], or recurrent networks [20, 21, 22, 23]. Moreover, a variety of design principles for ANNs have been developed [22, 24, 25] that can be used as building blocks in order to construct powerful models for new applications.

As an example, for an image classification task specific (very successful) designs have been proposed [26]. Each image is represented by a vector of pixel values that serves as input to an ANN. The neurons of the ANN are arranged in consecutive layers, where the first is the input layer and the last is the output layer. Neurons of one layer have incoming connections only from the neurons of the previous layer, and outgoing connections to the neurons of the subsequent layer (see Figure 2). The input-output function that is represented by the ANN is determined by the so-called *weights* that are attributed to the connections. The actual process of learning consists in the automated, algorithmic adaptation of these weights with respect to a training dataset. In DL, the most prominent learning algorithm is stochastic gradient descent [27] including its various variants. It works in repeated passes (epochs) over the training set and aims at iteratively adapting the weights in small steps upon every presentation of a batch of training examples. For ANN, the representation of an input is the activation of the neurons in a particular layer, where layers closer to the output layer have more abstract representation (see Figure 2). Typically, the first layer of the ANN will learn to detect simple local patterns such as edges and corners, whereas neurons in higher layers will become selective for more abstract prototypical patterns like an eye of a dog or a wheel of a car [28].

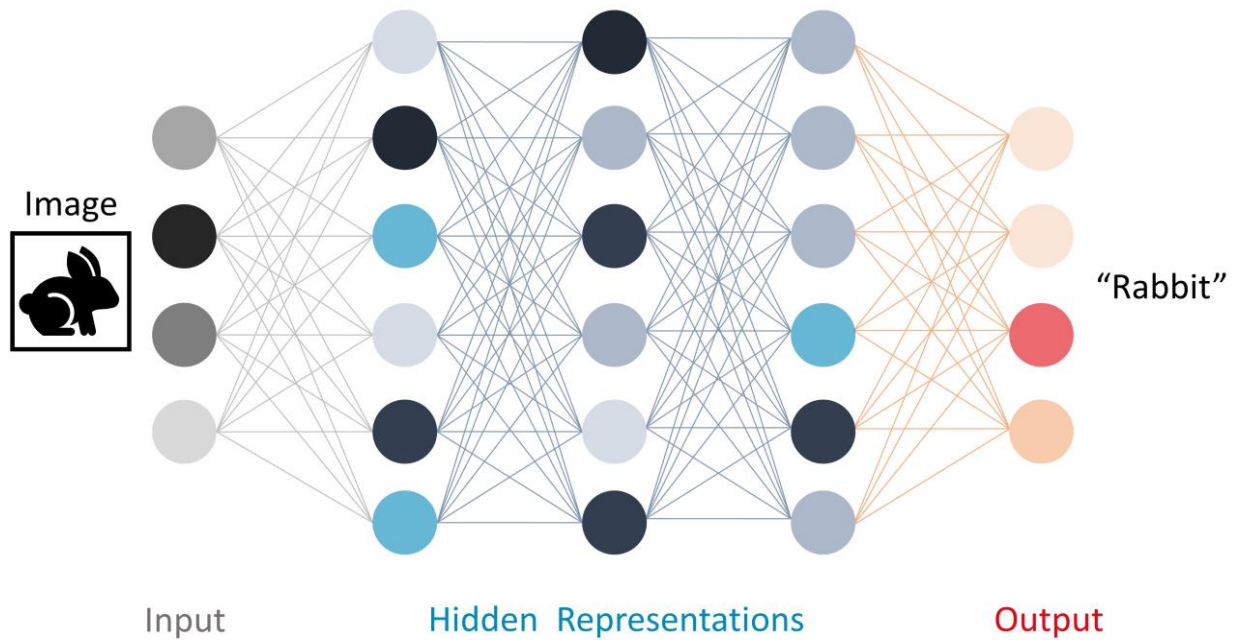

Figure 2: Illustrative example of a feed-forward network with multiple layers for image classification. From layer to layer the representation from the input image is transformed into a more and more abstract representation. Finally, a neuron in the output layer indicates the predicted class of the input image.

Figure 3 depicts the relationship between Artificial Intelligence, Machine Learning, and Deep Learning.

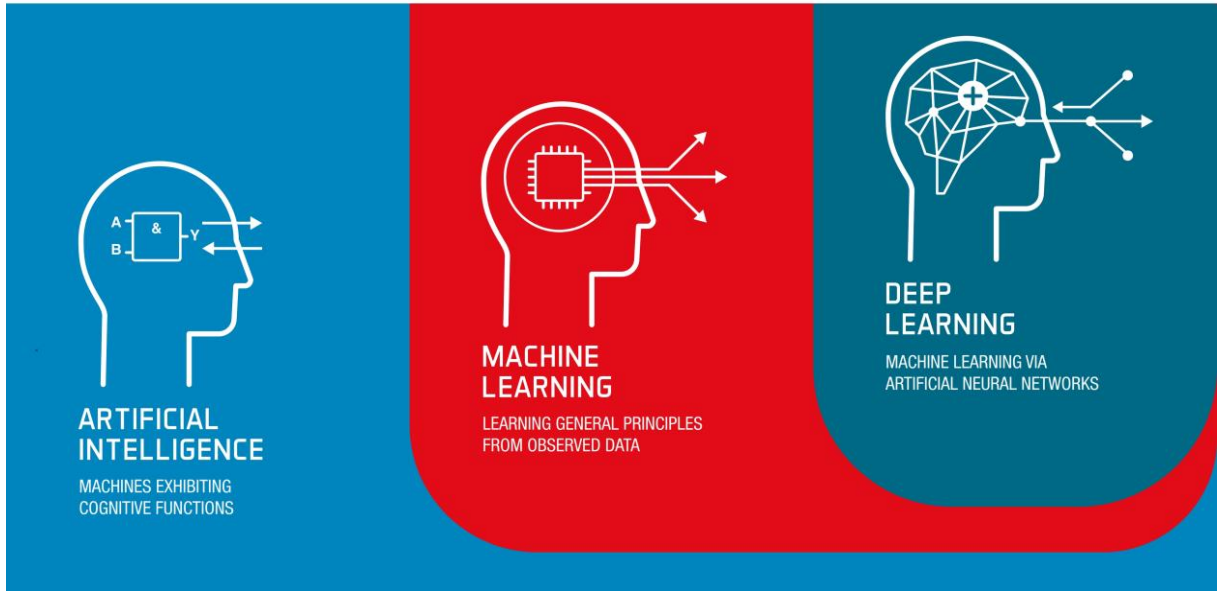

Figure 3: ML is a sub-category of AI while DL is a sub-category of ML. For example, expert systems are associated with AI, but not ML and kernel methods are associated with ML, but not DL.

In the following, we describe the most important learning types in ML.

**Supervised Learning:** Supervised learning is the ML task of learning a function that maps an input to an output based on a given training set of inputs and their targets, also called desired outputs or labels (see Figure 4 and Section 4.1) [29]. Supervised learning tasks can be categorized based on the nature of their targets:

- discrete values (classification)
- continuous values (regression)
- sequences (e.g., time series and texts)
- sets (e.g., point clouds)
- graphs (e.g., molecules and social networks)

Given a training set, supervised learning consists in selecting a specific model from a certain model class/hypothesis class. A model class can have parametrized models or parameter-free models. A typical example of a model class is an ANN with a specific architecture, where the connection weights are the free parameters of the (parametric) model class. A specific model is selected from this model class by selecting the parameters and then assigning them to the corresponding weights of the ANN. A supervised training algorithm uses the given training dataset for this model selection with the goal that the chosen model promises the best performance on future unseen data (the generalization performance). Such learning algorithms may comprise simple methods or complex and iterative methods. Simple methods may directly compute the parameters and may be as simple as computing the arithmetic mean (e.g., naive Bayes classifiers or linear regression). Complex methods include convex optimization (e.g., support vector machines or support vector regression) or gradient based stochastic iterative methods (e.g., ANNs and DL). Supervised learning methods assume that the input-output function as well as the data distribution is the same in the training set and the future unseen dataset. In most cases ML methods assume that both the training dataset and the future unseen dataset obeys the i.i.d. assumption which allows to estimate the expected future performance (the generalization error). The i.i.d. assumption states that there exists some stochastic process in the real world that produces data samples, one sample at a time, where the individual samples are statistically independent from each other. In simple words, both the training data and the future unseen data in the final application should stem from the same (static) real-world process. Of course, there are various methods that relax or alter these assumptions, like e.g., active learning. The performance of a trained model is usually evaluated on a test dataset that was not used – in fact not even seen by the developer – during model development.

Variations of supervised learning include weakly-supervised learning [30] and semi-supervised learning [31]. They are used if noisy, partial, or imprecise labels are given or when only a small portion of the data is labeled due to costly and/or time-consuming labelling processes. Another variation of supervised learning is self-supervised learning (e.g., next frame prediction), where labels are extracted efficiently in an automated fashion from the data itself. Often the production of new products or the acquisition of new customers generate data that has few training examples. In these cases, few-shot learning [32] and zero-shot learning [33, 34] enable learning with a small number of training examples.

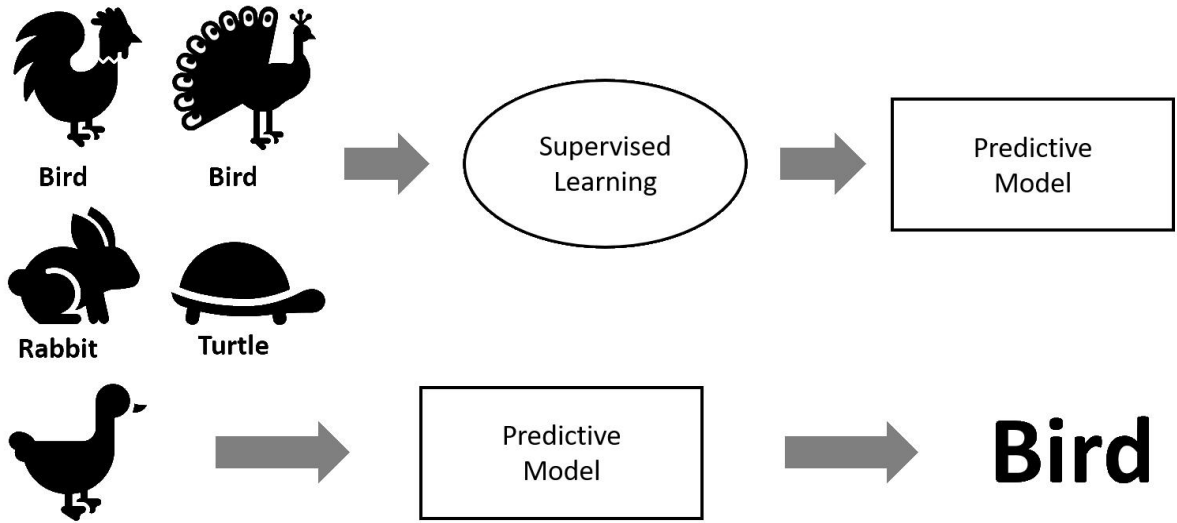

Figure 4: A predictive model is learned in a supervised fashion, using input-target tuples during training (top). The learned model can be applied to process new data in useful ways (bottom).

**Unsupervised Learning:** In contrast to supervised learning, unsupervised learning uses unlabeled data for training. Unsupervised methods try to extract structure in the data, represent the data in a more compact or more useful way, or build a model of the data generating process or parts thereof. In contrast to supervised problems, the quality of models in unsupervised problems is mostly measured on the cumulative output on all or a set of objects, which leads to a more complicated objective function than supervised learning. Therefore, credit assignment (contribution to the objective) to the processing of single data points and how it should be improved is more difficult.

While supervised methods are used for performing prediction of future data, unsupervised methods allow to explore the data, find structure in the data, visualize the data, or compress the data. Unsupervised methods can help to understand the data and to generate new knowledge. Unsupervised methods can be grouped into recoding methods and generative methods.

Recoding methods generate a new representation of objects given a representation of them as a feature vector. In most cases, they down-project or compress feature vectors of objects into a lower-dimensional space in order to remove redundancies and components which are not relevant. Projection methods comprise principal component analysis (PCA) [35, 36], independent component analysis (ICA) [37], factor analysis [38], or projection pursuit [39]. Cluster analysis (see Figure 5) is an important sub-field of recoding methods.

> "Cluster analysis or clustering is the task of grouping a set of objects in such a way that objects in the same group (called a cluster) are more similar or more closely connected (in some sense) to each other than to those in other groups (clusters)." [40]

Note that there is no general definition of similarity or connectedness, thus the precise aim has to be defined case by case depending on the application. Clustering is merely used as an analytic tool in exploratory data mining. Clustering methods include k-Means [41, 42], hierarchical clustering [43], mixture models [44], or self-organizing maps [45].

Generative models aim at modeling the data-generating process, i.e., they are a model of the real world, at least with respect to the specific application domain. From a mathematical point of view, the real world is a stochastic process that produces data samples according to a certain probability density distribution. Generative unsupervised learning tries to come up with a probabilistic model that describes that real-world distribution as precisely as possible, such that it is able to generate new artificial data with the same density. The aim is to obtain a world model for which the density of the data points produced by the model matches the observed data density. The data generation process may also have input components or random components, which drive the process. Such input or random components may be included into the model. Important for the generative approach is to include as much prior knowledge about the world or desired model properties into the model as possible in order to restrict the number of models which can explain the observed data. Generative unsupervised learning methods are density estimation (kernel density estimation [46, 47], Gaussian mixtures [44]), hidden Markov models [48], or belief networks [49]. The latter are subsumed into Markov networks or Markov random fields. Other and more advanced methods include restricted Boltzmann machines [50, 51], (modern) Hopfield networks [52], neural network auto-associators [53], variational autoencoders [54], generative adversarial networks [55], and normalizing flows [56].

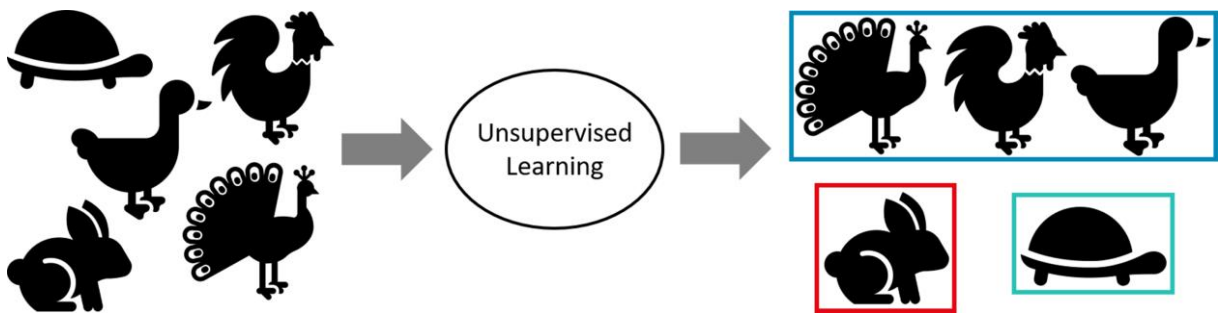

Figure 5: Clustering as an example of unsupervised learning. The structure in the data is analyzed and the objects are agglomerated in groups.

**Reinforcement Learning (RL):** It is another important learning paradigm, that significantly differs from supervised and unsupervised Learning.

> „Reinforcement learning is learning what to do – how to map situations to actions – so as to maximize the reward achieved by the acting agent. The learner is not told which actions to take, but instead must discover which actions yield the most reward by trying them. In the most interesting and challenging cases, actions may affect not only the immediate reward but also the next situation and, through that, all subsequent rewards. These two Characteristics – trial-and-error search and delayed reward – are the two most important distinguishing features of reinforcement learning" [57].

RL deals with learning via interactions with an environment (see Figure 6). An agent tries to maximize the accumulated reward that originates from these interactions. RL was the first method that produced interactive models that beat Go[1] masters.

---

[1] Go (game). In Wikipedia, The Free Encyclopedia. https://en.wikipedia.org/wiki/Go_(game)

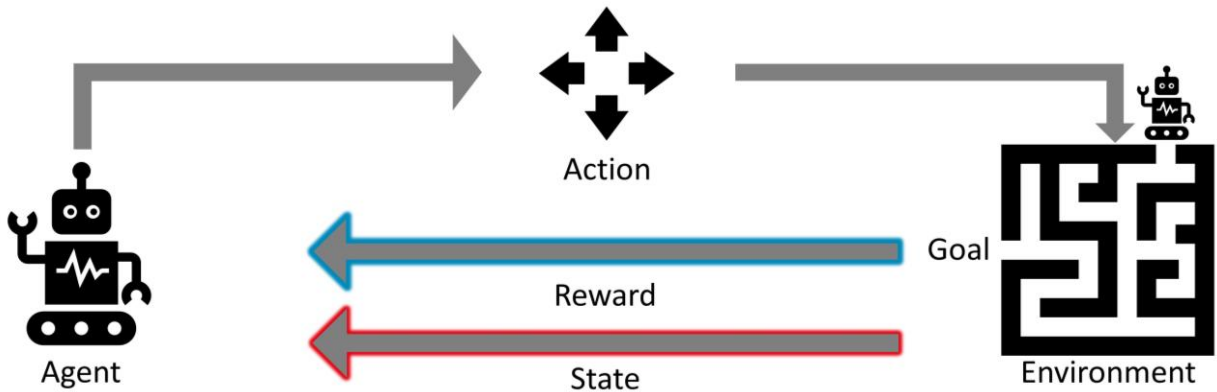

Figure 6: Reinforcement learning scheme. An agent interacts with an environment and learns strategies to maximize the accumulated reward.

## 2.2 Research Developments

Recently, modern AI has revolutionized various scientific fields of science and has started to improve or enable important commercial applications, which indicates AI's enormous potential. The recent surge in AI can be mostly attributed to advances in DL, where ANNs are trained on large datasets in order to solve complex tasks. For example, the outstanding results of DL in the *ImageNet Large Scale Visual Recognition Challenge* contributed to a revolution in the field of computer vision [26]. Deep convolutional networks [19, 18] led to breakthroughs in processing images and videos.

Moreover, DL revolutionized, speech recognition, natural language processing (NLP), text analysis, and entire fields, such as life science, automated driving, and the entertainment industry. Recurrent neural networks [22, 21, 20, 23] set a new state of the art for tasks that are related to sequential data, such as, text [58], and speech [59]. More recently, attention mechanisms [24] led to major contributions in NLP [60] and life science [61, 62].

DL evolved to an important framework for handling large amounts of data efficiently in terms of computation. The development of theoretically well-founded tools for DL and simultaneously ensuring high data quality are crucial to the success of DL in the near future.

Besides the recent advances with DL, non-parametric methods such as random forests and support vector machines (SVMs) are used heavily due to their simple interpretations. SVMs are even convex problems and have unique solutions. Both random forests and SVMs are especially superior in cases where no assumptions about the data can be made and only a moderate number of data points are available for learning.

In the context of certification, important research directions in ML include aspects such as stability [8, 63, 9] and data integrity [64, 10]. The stability aspect aims for obtaining tight bounds on the generalization error, whereas data integrity aims for ensuring a certain quality of data that is used to train ML models. The advent of adversarial attacks [65] revealed potential issues regarding safe and secure operation of ML models.

ML methods are more readily available today through open source implementations for popular programming languages such as Python[2] or R[3]. Popular libraries for these programming languages include TensorFlow[4], PyTorch[5], Scikit-learn[6], NumPy[7], and CUDA[8].

---

# 3  Standardization and Certification

In this section, we give an overview of current activities in standardization of AI, investigate reproducibility and explainability of ML methods, and list some major challenges with regard to certification of ML applications.

## 3.1  Standardization Committees and Groups

Standards are national and international unifications that have been formed through the establishment and implementation of common foundations for recurring applications within an interest group. The broad use of standards facilitates technological progress (e.g., simplified communication) and promotes the exchange of goods or services (e.g., compatibility, product safety). The number of AI applications is growing rapidly, which raises the need for standards. Such standards can support achieving important goals in terms of functionality, interoperability, as well as building reliance. The standards may include requirements, specifications, and guidelines that AI applications must fulfill in order to operate accurately, reliably, and safely [66].

Societal concerns and ethical aspects of AI are currently a core topic in AI standardization, since incorrectly developed or applied ML technologies can lead to potentially undesirable consequences. Moreover, ML technologies can be intentionally developed and used for unethical practices, which opens up another subject of debate. Considering the usage of ML-driven applications in safety-critical environments, it is furthermore mandatory that applied ML must not weaken functional safety and that certain safety processes are implemented in case of failures (fail-safe). Therefore, reliance is a necessary aspect to ensure that AI technologies are accepted by the public. Reliance is the fundament that allows a broad market introduction of AI systems. Norms and standards can provide guidelines for risk mitigation or provide best practices for the development of AI systems. In this context, certification plays a decisive role, as it builds on existing norms and standards and thus provides a basis for reliance on such complex technical systems. Thereby, reliance is generated by independent experts during a certification process, from the reputation of the accredited organization, and from the applied criteria and guidelines. With the help of certification, negative consequences can be avoided on the one hand, and developments in the interest of the common good can be promoted on the other hand. Furthermore, certification can lead to positive competitive dynamics through incentives for product improvement.

First steps towards AI standardization have already been taken (see Figure 7) on the international level[9][10] by the International Organization for Standardization (ISO), the International Electrotechnical Commission (IEC), the Institute of Electrical and Electronics Engineers (IEEE), the International Telecommunication Union (ITU), in Europe by the European Committee for Standardization (CEN), the European Committee for Electrotechnical Standardization (CENELEC), the European Telecommunications Standards Institute (ETSI), and on a national level among others in Germany by the German Institute for Standardization (Deutsches Institut für Normung e.V. (DIN)), and the German Commission for Electrical, Electronic and Information Technologies (Deutsche Kommission Elektrotechnik Elektronik Informationstechnik (DKE) im DIN und Verband der Elektrotechnik Elektronik und Informationstechnik (VDE)), and in Austria by the Austrian Standards International and the Austrian Association for Electronics (Österreichischer Verband für Elektrotechnik (OVE)).

---

[9] DKE. Basics of Standardization. https://www.dke.de/en/standards-and-specifications/basics-of-standardization
[10] DKE. The Importance of Standardization – Benefit and Advantages. https://www.dke.de/en/standards-and-specifications/importance-of-standardization

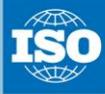

Figure 7: Overview of international and national standardization organizations.

In the following, the individual standardization committees and groups established at an international and European level are briefly described.

**ISO/IEC:** Since May 2018, the international *Joint Technical Committee 1* in *Subcommittee 42* (JTC 1/SC 42) of the ISO/IEC has been developing standards for the AI ecosystem, including all relevant and major topics concerning ML. In addition, guidelines for AI applications are also being developed. Figure 8 shows the work programs[11] and structure[12] of the JTC 1/SC 42 *Artificial Intelligence*.

**ITU:** The ITU-T Focus Group *Machine Learning for Future Networks including 5G* (FG ML5G)[13] was established in November 2017 and was active until July 2020. The focus group (FG) drafted ten technical specifications for ML for future networks, including interfaces, network architectures, protocols, algorithms and data formats. The ITU/WHO Focus Group on *Artificial Intelligence for Health* (FG-AI4H)[14] is working in partnership with the World Health Organization (WHO) to create a standardized assessment framework for the evaluation of AI-based methods for health, diagnosis, triage, or treatment decisions. It was founded in July 2018.

**IEEE:** The *IEEE Global Initiative*[15] has published a detailed work on the ethical consideration of automated and intelligent systems with its second version of *Ethically Aligned Design – A Vision for Prioritizing Human Well-being with Autonomous and Intelligent Systems* [67]. More concrete projects for the development of IEEE standards are currently being implemented in the IEEE P7000 series, which focuses on interoperability, functionality, and safety.

**CEN and CENELEC:** The CEN-CENELEC *Focus Group on Artificial Intelligence*[16], established in April 2019, addresses the need for AI standardization in Europe. In this context, the FG supports the activities of ISO/IEC JTC 1/SC 42. It seeks to identify specific European requirements and acts as an interface to the European Commission. The main objective of this FG was to develop an AI standardization roadmap for Europe.

---

[11] Details on JTC 1/SC: https://jtc1info.org/sd-2-history/jtc1-subcommittees/sc-42/
[12] Structure of JTC 1/SC 42: https://www.iso.org/committee/6794475.html
[13] Details on FG ML5G: https://www.itu.int/en/ITU-T/focusgroups/ml5g/Pages/default.aspx
[14] Details on FG-AI4H: https://www.itu.int/en/ITU-T/focusgroups/ai4h/Pages/default.aspx
[15] IEEE Global Initiative: https://standards.ieee.org/industry-connections/ec/autonomous-systems.html
[16] Details on CEN-CENELEC FG: https://www.cencenelec.eu/standards/Topics/ArtificialIntelligence/Pages/default.aspx

**ETSI:** There are a number of industry specification groups (ISGs) working in the AI/ML area, such as the ISG on *Experiential Networked Intelligence* (ENI)[17], which develops standards that use AI mechanisms to assist the management and orchestration of the network, the ISG *Zero-touch Network and Service Management* (ZSM)[18], which defines the AI/ML enablers in end-to-end service and network management, and the ISG on *Securing Artificial Intelligence* (SAI)[19], created in September 2019, which develops standards for securing AI from attacks, for mitigations against AI cyber-attacks, and for using AI to enhance security measures against attacks.

**DIN/DKE/VDE:** In Germany, standardization work on AI is carried out in the *DIN Standards Committee Information Technology and selected IT Applications, Working Committee Artificial Intelligence NA 043-01-42 AA*[20]. This committee develops standards and practices on tools, processes, and applications in the field of AI, considering societal opportunities and risks. The working committee essentially reflects the work of ISO/IEC/JTC 1/SC 42 *Artificial Intelligence* and the CEN/CENELEC *Focus Group on Artificial Intelligence*.

**Austrian Standards/OVE**: In Austria, standardization work on AI is carried out in the *Austrian Standard Committee 001 Information technology and its application, Working Group Artificial Intelligence AG 001 42*[21]. This working group elaborates the Austrian position in AI standardization and serves as a mirror committee to ISO/IEC JTC 1/SC 42.

At the moment there are several ongoing activities regarding standardization of AI (see Table 3 in the Appendix), including topics such as reliability and robustness, safety, ethics, fairness and non-discrimination, and human agency and oversight. In the USA, the FDA (US Food and Drug Administration) is currently working towards a regulatory framework for AI/ML-based medical software. China is currently working on their *China Standards 2035* plan, which aims to involve the country into the global standardization scheme. This plan includes a section on the standardization of AI. This variety of ongoing work demonstrates the need for standards and norms for AI.

---

[17] Details on ISG ENI: https://www.etsi.org/technologies/experiential-networked-intelligence
[18] Details on ISG ZSM: https://www.etsi.org/technologies/zero-touch-network-service-management
[19] Details on ISG SAI: https://www.etsi.org/technologies/securing-artificial-intelligence
[20] DIN AI working committee: https://www.din.de/de/mitwirken/normenausschuesse/nia/nationale-gremien/wdc-grem:din21:284801493
[21] Austrian Standard IT/AI working groups. https://www.austrian-standards.at/de/standardisierung/komitees-arbeitsgruppen/nationale-komitees/committees/1/details

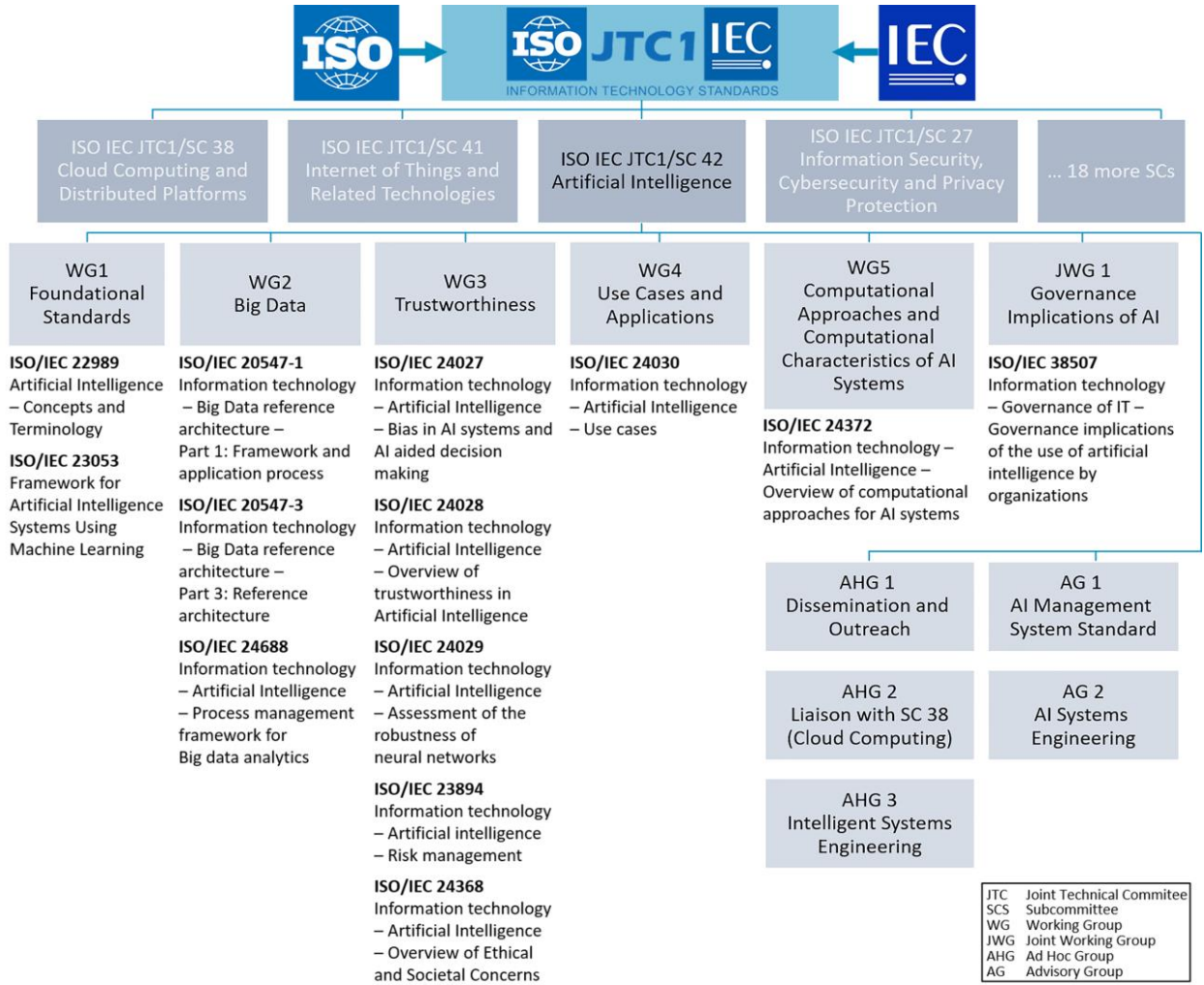

Figure 8: Overview of the work programs and structure of the ISO/IEC JTC 1/SC 42.

## 3.2 Explainability and White Boxes

In the public, ML methods, especially methods including ANNs, are often perceived as black boxes, which would mean that the internal working mechanisms would be inaccessible for anyone (including developers) and thus no knowledge about these mechanisms could be acquired. The prejudice goes that decisions driven by ML (regardless of the complexity of the learned function) were fundamentally incomprehensible and that ML systems can only be investigated via their input-output behavior. Thus, some people demand that ML methods should only be applied if the system is able to deliver convincing arguments together with the decisions it takes.

Note that the perception as a black box (or a white box as described below) depends on the context. For users of ML services these systems are in fact black boxes, since they typically do not have access to internal working mechanisms of ML models. However, this is not caused by the ML models directly, but by access restrictions due to reasons concerning safety and security.

However, in this paper, we would like to stay in the context of ML application development, where it is in fact possible to access internal working mechanisms of ML models. Thus, although ML models can be quite complex in practice, they are in fact white boxes (see Figure 9). An ML model is a combination of basic mathematical operations and all model-internal states and computations can be analyzed and traced down to every single bit and byte of its numerical operation. All computations are perfectly traceable and their results are unambiguously reproducible in terms of mathematical operations. This kind of reproducibility theoretically even applies to the training process that leads to the learned ML model. It is worth noting that reproducibility does not imply any kind of interpretability of an ML model as discussed, e.g., in the work of Zachary Lipton [68]. Thus, we intentionally do not include topics regarding interpretability in our notion of reproducibility, black boxes, and white boxes.

In practice, exact reproducibility would in principle be possible under well-posed conditions. For example, respective inputs to the ML system would have to be stored in order to reproduce the exact input-output mapping, including all intermediate calculations. However, in most cases such a degree of reproducibility is not needed for the purpose of certification. For example, reproducing a specific local minimum in the non-convex loss landscape of ANNs [69, 70, 71] would require fixing random seeds and considering numerical rounding errors. In our certification procedure, we do not require such a high degree of reproducibility. Instead, we rather demand for reproducibility (correctness, clear implementation, etc.) of the used ML algorithms that lead to models which meet some minimum performance requirements in a statistically significant fashion.

A real black box system would not allow such a degree of reproducibility for both how the model is selected (the training process) and how the learned model processes data. Moreover, a black box would not allow an analysis of model-internal states and computations. A black box could only be investigated via studying the model input and output. In order to be sure how a black box model behaves, one would need to study all possible input-output pairs of the model. For example, one could not apply continuity assumptions of the learned function without testing all possible inputs – which would be unfeasible for real-valued and/or high-dimensional inputs. A black box would negate any possibility to investigate and rate how the model processes data, which would in turn not allow for developing a certification procedure. Keeping this in mind, we note that complex model classes or complex learned functions should not be confused with black boxes.

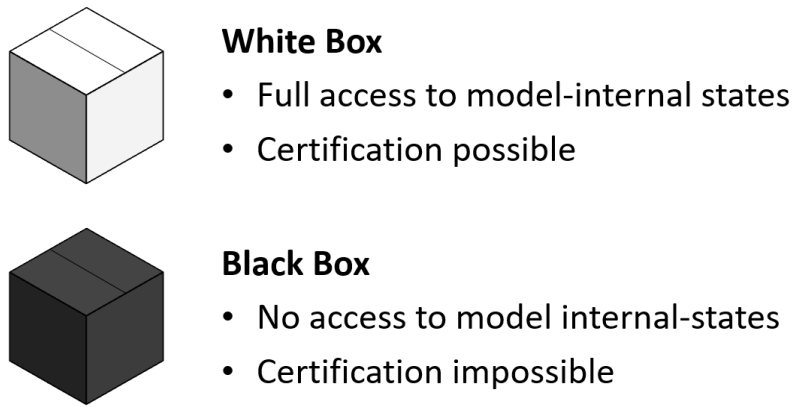

**White Box**
- Full access to model-internal states
- Certification possible

**Black Box**
- No access to model internal-states
- Certification impossible

Figure 9: ML models are white boxes, which allows for developing a certification procedure.

Although ML models are white boxes, it is true that both the high complexity of the model class and the sheer number of operations makes it very difficult or even impossible for a human to understand the process as a whole. Humans tend to like simple explanations for decisions and are often willing to offer simple arguments for their own decisions if asked. But in fact, it was shown that most human decisions are the result of the complex integration of life-long experience based on thousands of pieces of information [72, 73]. The simple arguments that humans can give in complex decisions are merely a-posteriori justifications or eventually a pedagogical tool for teachers but no real underlying cause for the brain-internal decision making. Finding arguments that convince humans to accept a taken decision is another problem by itself that should not be confused with the first problem of taking the decision. Moreover, human history has demonstrated that even convincing arguments need by far not to be true. This could also be the case for ML systems that provide additional explanations for taken decisions. Thus, requiring ML to justify its decisions with seemingly convincing arguments seems like the wrong way to go on the long-term, until it is possible to develop a methodology which ensures that the given explanation correctly represents the chain of causality.

Arguments given for pedagogical reasons are precisely the simplifications and verbal abstractions that enable humans to teach and learn valuable knowledge about the real world. Thus, trying to distill a trained ML model and extract knowledge from it is a promising line of research by itself. Correlations between the model's input and output (including internal representations) may be studied via interpretability methods such as LIME [74], SHAP [75], or integrated gradients [76]. Moreover, visualization techniques may allow us to analyze high-dimensional data [77, 78] and even understand hidden representations of certain model classes [28, 79]. These methods allow for investigating and understanding the bigger picture of the abstractions and working mechanisms of trained, highly complex ML models and might also be suitable instruments for building reliance and raising public acceptance for ML systems.

## 3.3 Challenges Regarding Certification of Machine Learning Applications

Supervised learning in the context of ML is one of the most influential technologies in recent years. Due to its high flexibility and efficient algorithms, many ML models that solve real-world problems are trained in a supervised fashion. Thus, the aim is both to ensure a certain quality of the ML system and to raise reliance and acceptance for such systems in the public in general. The problem of supervised learning has a clear definition and a long scientific history, dating back to the very beginning of ML in the last century. Because the problem definition is clearly specified, it is possible to come up with well-defined rules that evaluate and rate specific methods, algorithms and realizations that emerge from this framework. These provably correct mathematical rules allow for developing a certification process that ensures a certain level of quality and reliability for the algorithm under test.

Although trained models are typically quite complex, all aspects of data processing, learning algorithm, and the learned model can be reproduced. Since the trained model is a deterministic mathematical function, the same inputs will always lead to the same outputs.

However, there are a number of ML methods that are commonly used in current applications that go beyond the scope of the supervised learning framework and thus exceed the scope of the quality assessment approach that we apply for certification. RL is such an example. In RL, an agent interacts with an environment and aims to maximize the accumulated reward. RL is extensively used in AlphaGo, which allegedly produced the best Go player on earth.

The game Go, regardless of its high complexity, is a deterministic game and optimal moves could be computed in theory. However, for the sake of simplicity of the following thought experiment, we assume that optimal moves cannot be computed. This assumption often applies to real-world tasks and to tasks with continuous state- and action spaces.

The AlphaGo agent learns how to play Go, it learns to make moves in order to maximize the chance of winning the game. In contrast to supervised learning, where we know at least the targets for some examples, in RL applications we do not know targets for the individual moves, but only the reward, from which the RL system derives the most suitable moves as the solution of the problem. It is impossible to predict a-priori, i.e., before the training process has started, which strategies the agent will come up with and which individual moves the agent will generate as outputs. And even though it is clearly deterministic which move the agent will take next, this cannot be predicted from outside without introspecting the inner states of the agent (white box). In all these struggles for guarantees and certifications, it is necessary to clearly realize that it is fundamentally impossible to answer the question whether the trained AlphaGo agent plays the correct move, which by definition would have to be the best possible move given the game situation.

In such cases we might be tempted to ask, if we could demand for some minimal guarantees at least. For example, assuming we had a rule-based definition of "silly" mistakes, we could aim to certify that the Go playing agent does not make such mistakes. This turns out to be a fallacy if our aim was to have an agent that maximizes the chance of winning games. A seemingly silly move may turn out as the ingenious solution to win the game in the end. Typically, this is a move that just no other player would have thought of or would have dared to make [80]. By forcing the agent to stick to some predefined rules, just because we want to have something that we can guarantee, we would have hindered the agent from taking that best ingenious move and our agent would perhaps have lost that game. This should make it clear that it is not only fundamentally impossible to certify the correctness of each individual decision of such a system, but even worse: enforcing guarantees that are not perfectly aligned with the optimization goal always reduces the performance with respect to the true underlying goal. Thus, RL setups will require different approaches in order to get certifiable performance measures, which goes beyond the scope of this work.

Another problem arises when trying to align human values with objectives of ML algorithms. Real-world ML applications, especially when applied at large scale, can lead to unexpected effects and problems within the frame of the alignment problem [29, 81]. Although optimization objectives typically lead to the desired model behavior, a truly intelligent agent could figure out alternative ways to achieve its optimization objective while completely ignoring human values. A prominent thought experiment of the alignment problem is known as the paper clip problem, where an agent tries to produce paper clips at all cost. A real-world example of the alignment problem can be observed in the field of marketing, where an ML agent is trained to maximize user engagement on a social media platform. The objective leads to learning to recommend radical political content, and successively to political radicalization of users [82].

In practice, we observe that highly complex ML approaches typically raise challenges when it comes to developing a certification procedure. Thus, in the following we list some important challenges that need to be investigated in order to meet the requirements for a reliable quality assessment of ML applications in the future. This holds true especially for industrial applications.

1. **Theoretical understanding of ML key concepts is essential:** Often, ML developers find themselves in situations where it is difficult to check and verify whether mathematical assumptions hold for a given practical problem. The problem at hand or the applied method thus may extend beyond the ground of the strong mathematical theory. A silent but unchecked assumption of the training data and the future unseen data being independent and identically distributed (i.i.d. assumption) is the most frequently encountered example of such a fallacy. Note that while highly qualified developers may be well aware of these issues, we also aim to address less qualified developers when it comes to product certification. Thus, a core challenge is correctly interpreting and checking mathematical prerequisites for specific realizations in ML applications.

2. **Lack of quality assessment:** Currently, many ML models are deployed without appropriate quality assessment. This leads to erroneous applications and promotes potential security vulnerabilities. Therefore, standardized quality assessment procedures must be established. Again, this holds especially true for potentially less qualified or unexperienced organizations.

3. **Handling of domain gaps:** An unsolved problem regarding certification arises from domain gaps (also called distributional shifts) [83, 84, 85, 6]. Domain gaps arise when the data distributions for learning and inference do not match, which can be quite common for real-world applications. A typical scenario is learning from simulated data and applying the model in the real world with slightly different data [86]. Another prominent example can be found in automated driving, where data is typically gathered in the USA or Europe, but trained models are then deployed in other regions.

4. **Fast advancements in ML technology:** Due to the large amounts of data and the availability of better hardware, the field of ML has experienced fast developments during recent years. The fast progress of scientific discoveries and the pressure to immediately use them in practical applications (or in scientific work) can lead to an increased risk of reduced quality of the resulting ML systems. This pressure often prevents developers from thoroughly understanding the underlying assumptions, implications, and proofs that apply to their applications.

5. **Lack of clear requirements:** In practice, many ML applications can fail due to wrong usage, either via inexperienced developers or via the users. To reduce the problem of wrong usage, there is a need for extensive and clear communication about the task to be solved (requirements), the details about the implemented methods, and their limitations.

6. **Lack of confidence measure:** Most ML models do not offer a justified confidence measure of the model's uncertainties. For example, in classification models, the probability vector obtained in the top layer (predominantly softmax output) is often interpreted as model confidence. However, functions like softmax can result in extrapolations with unjustified high confidence for data points far from the training data, hence providing a false sense of safety [87].

7. **Ethical considerations:** In accordance with the high European standards and laws, a comprehensive certification process should not only include technical aspects of the ML system, but also ethical ones. Consequently, the requirement is to be aware of the ethical dimension of the overall decision-making process, especially in terms of gender and racial discrimination and in terms of environmental impact. As an example, we might think of a system that is trained to guess or predict the current salary of a person given its curriculum vitae and various other additional information. It is thus of utmost importance to clearly distinguish between prediction models that aim for reflecting the reality in an unbiased way and decision models that aim for selecting an appropriate choice with respect to a certain goal definition, i.e., optimize the decision process regarding the goal. Ethical considerations must be considered explicitly in designing the goal definitions and should never be implemented by an artificial distortion of input data or an artificial inductive bias of the ML model. The question if the goal definition itself is within the scope of the law and ethical guidelines is a question for lawyers and courts and beyond the scope of technical certification of ML systems.

8. **Raising acceptance:** ML systems are often claimed to contain non-reproducible and/or non-interpretable components that would raise serious safety issues. A key challenge is communicating the fundamental principles of ML in easy ways in order to raise acceptance and reliance on such systems for both the public and customers.

9. **Robustness against attacks:** Adversarial attacks [65] pose a threat to the safety of ML applications. Such attacks are intentionally designed and aim to manipulate the inputs of ML models in order to force them to differ from expected functioning. Often these inputs are manipulated only by a tiny degree and can be indistinguishable from a valid input when inspected by human experts, yet they cause entirely unexpected model outputs [88, 89]. This can lead to critical failures and hazardous situations. In addition to the ML aspect, classical software development security issues also play a major role in the context of attacks.

10. **Lack of qualified personnel:** Since many important theoretical concepts about ML are unknown to the ML community in practice, there is a need for experts who are able to develop, analyze, and evaluate ML applications. As ML is already widely used in industry, it is essential to ensure that appropriate investments in ML education are made, both in the short and long term.

11. **Designing flexible certification pipelines:** The large diversity and amount of different approaches in ML demand for flexible certification pipelines that are easy to modify, for example through self-consistent, modular components. Interacting elements of the ML system (data, model class selection, optimization objective, learning algorithm, etc.) require inspections from different points of view, which has to be treated coherently in a certification process. Moreover, many research areas in ML such as uncertainty estimation, interpretability methods, anomaly detection, adversarial attacks/defenses, and interaction with humans or environments are still active areas of research. Because these areas can potentially contribute to the development of more coherent certification processes, new findings need to be incorporated into existing certification pipelines promptly to tackle future challenges.

Based on the future challenges mentioned above, by introducing an audit catalog for ML we aim to lay the foundation for the certification of ML applications. At the moment, we restrict ourselves to certify low-risk applications and models that are trained in a supervised fashion. With low risk, we refer to applications that have a criticality level (CL) of 1 or 2. The notion of CLs is introduced in Section 4.2. The audit catalog covers all major aspects of supervised learning, including data processing, model selection, and the learning algorithm. The catalog allows for analyzing and rating all major components of the ML system in a systematic manner. We especially aim to include highly complex ML model classes that allow for solving complicated real-world problems.

# 4 The Audit Catalog for Machine Learning Applications

In this section, we propose an audit catalog to take the next step towards certification of ML algorithms. We believe that standards, norms, and regulatory frameworks will be particularly important for a healthy ML ecosystem, which includes developers, products, as well as customers. We aim to build public reliance in ML systems by ensuring a certain level of quality of the underlying algorithms. The following chapters describe the scope and the different aspects of the audit catalog, outline the general workflow of the certification process, highlight the design and structure of the catalog, and demonstrate its usage via selected scenarios.

The audit catalog consists of a requirements part and an auditor's instruction part. As usual in standards, the requirements part will be publicly available. The actual requirements catalog or more detailed information on the topic of certification of ML applications (Trusted AI) can be requested by email to the TÜV AUSTRIA Group: digitalservices@tuv.at

## 4.1 Scope

The audit catalog can be applied to ML systems within the supervised learning setting, which is one of the major drivers behind recent ML breakthroughs. At the moment, the audit catalog is not applicable to reinforcement learning, unsupervised learning, and other advanced ML techniques such as meta learning or active learning.

As mentioned in Section 3.2, ML models are in principle white boxes due to the possible access of model-internal states and computations. However, white box testing for ML cannot be directly compared to white box testing for classical software. While for classical software, every line of source code is analyzed, in ML one analyzes both data and algorithms that result in the final program - a trained ML model. Thus, our certification process can be regarded as a white box testing approach in the sense that we validate the data and algorithms that lead to a program rather than trying to interpret the program itself, which would be non-constructive.

Since it is impossible to verify correctness formally via mathematical proofs, our audit approach aims to validate whether an ML approach is reasonable, correct, meaningful, and clear. For example, our approach does not require a mathematical proof why a specific object has been detected in a given image, but rather verifies the data and ML algorithms that lead to the learned function. Moreover, for an audit it is often more constructive to check for the existence of a certain process rather than verifying the correctness of that process.

One important component of our audit catalog deals with the evaluation of the system using a representative test dataset. Without ground-truth test data, the risk of unexpected situations cannot be investigated properly. Additionally, we require the data to have several properties that are essential for ML. For example, we assume that all data distributions (training-, evaluation-, testing-, and new data) match and we assume that the distribution of new data does not change once the learned ML model is deployed. As also discussed in Section 3.3, these assumptions might not hold for high-risk applications such as automated driving, where data is obtained for one region, but the learned model is deployed in other regions (domain gaps). Although these high-risk applications are of particular importance for some fields, we do not investigate them extensively in this work. We rather follow a bottom-up approach and focus on low-risk applications (CL 1 or 2) in order to prepare our certification process to be extended to include high-risk applications (CL 3 or 4) in the future. Moreover, we believe that at the moment, applications with CLs of 1 and 2 will be more common in the industry than CLs of 3 or 4, further supporting our bottom-up approach.

A learned model is certified and must not be changed or altered afterwards. Updates or small changes to the system have to be recertified with additional follow-up audits. Typically, a recertification is less extensive and can be executed more quickly compared to a complete certification. Recertifications may thus be relevant for ML applications that are updated on a regular basis.

As mentioned above, the ML application has to be within the scope of supervised learning. In supervised learning, a dataset $D$ is defined as a set of $n$ labelled data points:

$$D = \{S_1, \ldots, S_n\}$$

Every data point $S_i$ can be written as a tuple $(x_i, z_i)$, where $x_i$ is a specific input and is represented by a feature vector. $z_i$ represents the corresponding target (label). Common data types for $x$ and $z$ include discrete values (e.g., class labels), continuous values (e.g., pixel intensities), sequences (e.g., human speech), sets (e.g., lidar point clouds), and graphs (e.g., molecules). In supervised learning, one seeks to find a function.

$f: X \to Z$, being an element of a model class $F$ of possible functions (e.g., a neural network architecture). For model development, the dataset is split into subsets according to the underlying theoretical framework of empirical risk minimization. A possible approach includes splitting the data into sets for training, validation, and testing. Alternatively, cross-validation, including variants such as cluster cross-validation [90] may be used.

The aim is to learn a model $f$ that approximates the target values $z_i$ from the inputs $x_i$:

$$z_i \approx y_i = f(x_i, \Phi)$$

where $y_i$ corresponds to the model output when given the input $x_i$ and (learnable) model parameters $\Phi$. Hyperparameters (e.g., model architecture, learning rate) are selected and adjusted for the given task and dataset. The free model parameters $\Phi$ are selected (learned) via an optimization algorithm like stochastic gradient descent, Hebbian learning, or other search algorithms. The optimization objective depends on both the model outputs and the corresponding targets:

$$\min_{\Phi} \sum_{i=1}^{N} loss(y_i, z_i)$$

where *loss* is the loss function between a model prediction $y_i$ and a target $z_i$. Common loss functions include distance measures such as the squared distance for regression problems and the negative cross-entropy for classification problems. The training procedure seeks to minimize the total loss on the training dataset by adjusting the model parameters $\Phi$. Note that optimization objectives are most often differentiable, allowing for calculating error gradients with respect to $\Phi$, which can then be used to improve in an iterative training process. The learned function $f$ is said to generalize if it performs well on unseen data. The quality of a model is evaluated using an evaluation metric. Ideally, the chosen loss function for training should result in a model with a good evaluation performance. Note that evaluation metrics are typically non-differentiable. We provide descriptions of some common loss functions and evaluation metrics in the Appendix.

## 4.2 Criticality Levels

In order to be able to map the criticality of an ML application regarding the effects of its decisions on people, environment, and organizations, we introduce our concept of criticality levels (CLs), which are inspired by the criticality pyramid of the Data Ethics Commission [91] and by the glossary of the Committee on National Security Systems [92] (see Table 1). As a result, the degree and level of detail for testing an ML application depends on the desired CL, which is an essential part of our certification pipeline. Each point in the audit catalog has an associated CL. In order to gain a certification at a specific CL, every point with both the same level and with potential lower levels must be evaluated and fulfilled. The CL is determined during the certification preparation process and the certification is conducted according to the defined CL. We define four different CLs in total.

The impact potential is the severity of consequences in case the ML system fails its assured properties. We define four levels of impact potential (see Table 1). Each level is associated with a set of requirements on the ML application. The levels indicate the potential consequences along several dimensions, which can be harm to life or limb, confidentiality of data, privacy of information, harm to the environment, ethical concerns, or other impacts. Thus, we do not consider the list to be complete. The highest impact potential on any dimension determines the overall CL of the ML system.

At the moment, the catalog can be applied up to a CL of 2. As many aspects for higher CLs require extensive research and field experience, the catalog will be constantly extended and refined to include higher CLs in the future.

Table 1: Criticality levels for certification of ML applications are identified through the highest impact potential of the application.

| CL | Impact Potential (Examples) | ML Application Requirements |
|---|---|---|
| 1 | No risk of harm to living beings, no risk of loss of confidential data, no ethical, or privacy concerns. | Basic minimum requirements of a competently developed ML application are fulfilled. |
| 2 | Living beings could be harmed with limited, non-permanent damage. Temporarily unavailability of non-critical data and services, violation of ethical concerns without identifiable harm to actual persons. | The ML application is developed according to industry standards and follows best practices that are regarded as state of the art. |
| 3 | Living beings could die or be restricted for life; the environment could be damaged. Manipulation of data with severe financial consequences, loss of control of the system to malicious attackers. | The ML application is developed and documented with great care. Safety & Security is ensured with processes and techniques that go beyond traditional best practices and industry standards. |
| 4 | Many living beings could die or could be restricted for life; the environment could be damaged permanently. Loss of information which endangers the existence of the organization. Long-term unavailability of critical data or services without which the organization cannot function. | The ML application is developed and documented with great care. Safety & Security is ensured with processes and techniques that go beyond traditional best practices and industry standards. All components of the ML application are formally secured and validated. |

## 4.3 Certification Procedure

In the very first step of the certification process, the organization to be audited receives a requirements catalog which contains a description for the individual requirements from the audit catalog. The requirements catalog is based on the audit catalog in terms of main chapters and sections but explains the purpose and objective of the individual subject areas in detail. However, the full document of the audit catalog, which includes additional information such as further proof requirements and notes, remains private to the auditor. In Table 2 we present an excerpt of our audit catalog.

The certification process for ML applications typically starts with a gap analysis and the preparations based on the requirements catalog, which is performed by the developers of the application (see Figure 10). The purpose of the gap analysis is to obtain an overview whether all necessary safety and security requirements have been considered in design and development of the application, whether the secure development processes have been documented adequately, and whether there is evidence for fulfilling the functional requirements regarding ethical aspects, data protection, etc., to handle possible issues early.

The next step is a kick-off meeting between auditors and developers and/or ML application owners, where the scope of the certification is precisely defined. Subsequently, the existing documentation is evaluated and reviewed by the auditors regarding the adequacy of process descriptions, specifications, regulation of responsibilities, and compliance with the requirements of the chosen CL. This is followed by audits in the form of interviews with the persons who are responsible for the specific processes in the organization to be audited.

After the interviews, a technical inspection based on the functional requirements is performed. During this inspection, records such as meeting logs, test reports, monitoring program documentation, and measurement results are checked.

The results of the audits are then presented in detail in a written report and an oral presentation. The audit findings contain all checkpoints, both the positive findings and the negative findings. The negative findings are split up into substantial and non-substantial nonconformities. In addition, the audit report states the subjects of examination, how these subjects were checked (test procedure), and how the operational effectiveness of the ML application has been tracked. If the audits are accomplished successfully, a certificate is issued. Thereafter, a monitoring audit takes place every year or after major changes of the application. The certificate is valid for three years, afterwards a recertification is necessary.

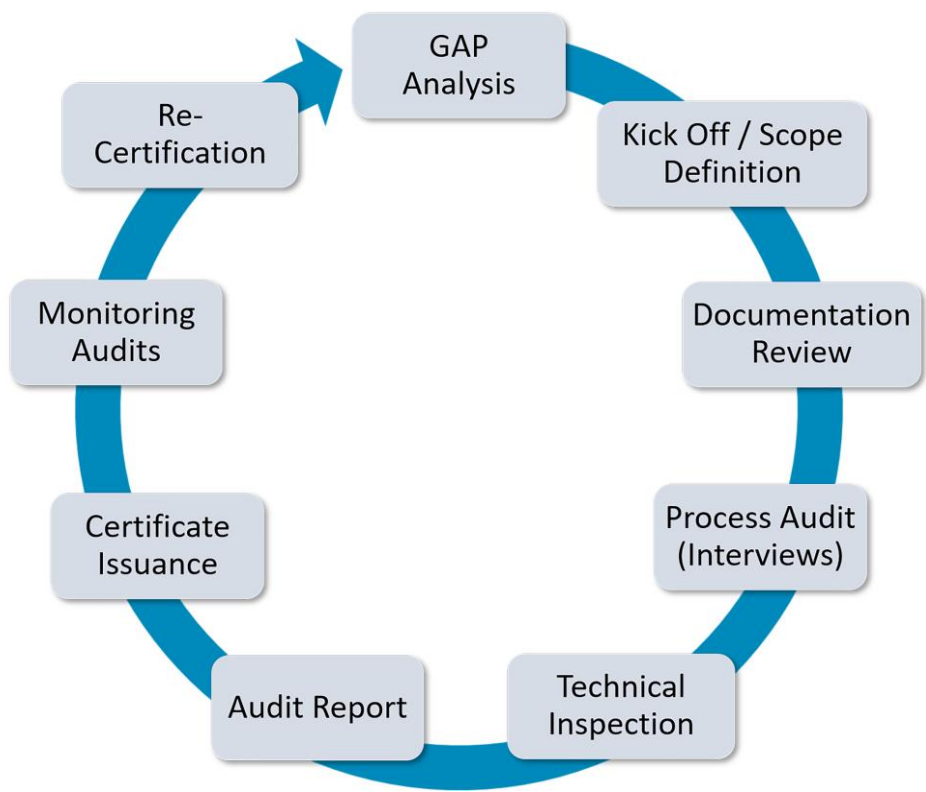

Figure 10: Workflow of the certification process for ML applications.

## 4.4 The Audit Catalog – Overview

The audit catalog is divided into three main chapters, comprising 33 sections with a total of approximately 200 requirements. Special attention is paid to the main chapter *Functional Requirements* and its subsection *Model Selection*, which are specifically addressed subsequently. The chapter *Functional Requirements* is most relevant for ML-related topics. Lastly, the audit catalog is applied to selected pitfall-scenarios. We demonstrate how the catalog can be used to reveal, investigate, and avoid such pitfalls. The main chapters of the audit catalog are listed below:

**Security in Software Development:** Since every ML application is first and foremost a software application, many aspects of the TÜV AUSTRIA audit catalog for secure software development apply to ML applications as well. These include topics such as regular training for developers, testing, patch management, correct use of containers, and secure deployment. In total, this main chapter contains 21 sections: *Awareness, Safety and Business Impact – Risk Assessment, Training, Specification, Design, Concepts, Technical Procedures, Environment, Cloud and Third-party Sources, Technical Stability and Security, Accountability, Implementation, Security Testing, Deployment, Patch Management, Quality and Integrity of the Data, Security Response, Security Metrics, Agility, Container,* and *Documentation*.

**Functional Requirements:** The heart of the audit catalog are the *Functional Requirements*. This main chapter covers all topics concerning model development, including data, methodology, model selection, and documentation. In total, this main chapter contains 12 sections: *Business Case, Data Collection, Data Preprocessing, Data Analysis, Model Selection, Model Requirements, Qualitative Model Inspection, Model Deployment, Operation, Failure Handling, Documentation and Communication, and Explainability and Interpretability*.

**Ethics and Data Protection:** In the case where ML has been used in conjunction with any kind of personal data or with direct impact on natural persons, ethics is an important aspect to consider when certifying an ML application. This main chapter of the audit catalog ensures that potential societal biases have been considered and resolved. Data privacy is also an important part of any software application and as such also part of our certification process for ML applications. It is ensured that personal data is handled correctly according to the General Data Protection Regulation[22] (GDPR) standard. In total, the Ethics and Data Protection main chapter contains 9 sections on ethics and 14 sections on data protection. The ethics sections include *Impact on Fundamental Rights, Priority of Human Action, Human Supervision, Transparency - Communication, Avoidance of Objectively Unjustified Discrimination, Accessibility, Consideration of Stakeholders, Minimization and Reporting of Negative Effects,* and *Codes of Conduct* while the data protection sections include *Privacy by Design / Privacy by Default, Quality and Integrity of Data, Data Storage, Authorization Concept, Data Minimization, Accuracy, Storage Limitation, Controls, Obligations of Confidentiality, Data Subject Rights, Automated Decisions on a Case-by-case Basis Including Profiling, Employee Training, Examination DPA,* and *Detection of Data Breaches.*

## 4.5  The Audit Catalog – Chapter "Functional Requirements"

The main chapter *Functional Requirements* of our audit catalog deals with all important ML-related aspects, including topics such as data, methodology, model selection, and documentation. The chapter is split up into several sections:

**Business Case:** The business case of the ML application and the business context needs to be clearly defined. The performance requirements of the ML application must be well defined, meaningful, and documented. The roles (e.g., project supervisors, software developers, domain-experts) in the project are defined and it is evident that the team has the competence to carry out the project.

**Data Collection:** The underlying data-generating process is an important factor for the development of a sound ML application. Thus, the data-generating process must be well documented. This also applies to data obtained from a third party and to data that is used for pre-training.

**Data Preprocessing:** The data preprocessing pipeline is an essential part of any model development and needs to be checked for correctness and consistency. Wrong preprocessing can potentially violate mathematical assumptions, which can lead to a variety of unexpected negative consequences such as data leakage. Data preprocessing includes all operations that are performed on the data, such as outlier detection and removal, data cleaning, data splits, augmentation during training, etc. This also applies to data that is used for pre-training.

**Data Analysis:** The data has to be representative considering the given task. Moreover, the data has to be investigated for potential biases and inconsistencies. This also applies to data that is used for pre-training.

**Model Selection:** This section of the audit catalog ensures that the ML model is developed using scientifically established methods and according to ML theory. This includes both model class selection, model selection (training), and correct methodology.

**Model Requirements:** This section covers the requirements for the model and model development. The model class, optimization objective(s), evaluation metric(s) and the given task have to fit together in a consistent way. Minimum performance requirements must be defined as a measure of success for a given task. Moreover, the model development must be reproducible. If necessary, model uncertainty has to be investigated by the developers.

**Qualitative Model Inspection:** This section ensures that all necessary steps are taken to verify and validate the quality of the learned model. This includes, e.g., checking for invariance to perturbations, stress tests, detecting and resolving learned (gender, racial, political, religious, etc.) biases, analysis of edge cases, and adversarial attacks.

**Model Deployment:** In case that a model that is developed on system A but deployed on system B, there have to be guarantees (e.g., by the provider of system B) that the executed deterministic function of the model on system B is identical to the function on system A, which was quality-checked. Well-established exchange formats like ONNX or PMML may be helpful to port models across libraries and languages, but due to subtle functional differences in the implementation of ML functions a superficial similarity of the systems should not be trusted.

**Operation:** Correct operation must be ensured at any time. During operation we assume that the data distribution remains quasi-stationary, at least until recertification. The system has to be tested and recertified periodically.

**Failure Handling:** In case of a failure, it is essential that certain safety processes are implemented (e.g., failover or switchover). This applies to all major components of the system, including both hardware and software. For real-time applications, a monitoring system should be used.

---

[22] General Data Protection Regulation (GDPR). https://gdprinfo.eu/

**Documentation and Communication:** The application needs to be fully documented. A detailed technical documentation for developers and domain experts (e.g., for code maintenance issues), as well as a general user manual must be provided.

**Explainability and Interpretability:** For higher CLs (3 and 4), where the usage of the ML application is more safety-critical as compared to lower CLs, it is important to be able to interpret the input-output mapping to some degree. This interpretability should have no negative impact on model performance.

## 4.6 The Audit Catalog – Section "Model Selection"

In the following, we present the section *Model Selection* from the main chapter *Functional Requirements* of the audit catalog. Table 2 shows how the individual entries are formulated in detail. Each topic has an associated CL. In order to receive a certification at a certain CL, all requirements of the corresponding and potential lower levels must be evaluated and fulfilled. Some critical entries within the audit catalog must be fulfilled entirely in order to acquire a certification (marked with "crit."). Most entries contain some additional information for auditors. However, this information is confidential and thus not presented in this white paper.

Table 2: The section *Model Selection* as part of the main chapter *Functional Requirements* of the audit catalog. Some entries are mandatory (marked with "crit.") and must be fulfilled entirely to acquire a certification according to the given CL.

| CL | Topic | Requirement Description | Proof Requirements |
|---|---|---|---|
| 1 | ML approach | A supervised learning approach is well-motivated and necessary. | It is impossible to solve the task sufficiently well (see requirements) using non-ML methods. Plausible reasons are given why a supervised learning approach is necessary and appropriate to solve the given task. |
| 1, crit. | Training and Validation datasets | An appropriate data split method is implemented according to the framework of empirical risk minimization and the given task. | For any data split method: All splits / folds meet the requirements (see Data Collection - Data is representative). There is no data leakage between splits / folds. Splits / folds are obtained, e.g., via random or temporal selection of data points or via custom approaches that are justified. For Cross-validation: The number of cross-validation folds is justified according to the given data and task (see Business Case - Requirements Specification). For Cluster Cross-validation: clusters in the data are investigated properly. Specific clusters are assigned to specific folds in a meaningful way under consideration of the given data and task. |
| 1, crit. | Test dataset | In addition to the training and validation splits, a test split is used. The test set performance is evaluated after the model development phase. | The test set is representative (see Data Collection - Data is representative). It is ensured that the test set was not used during the model development phase. It is ensured that the test set performance is evaluated after the model development phase. The test set performance meets the defined minimum performance requirements. |

| CL | Topic | Requirement Description | Proof Requirements |
|---|---|---|---|
| 2, crit. | Field test | The model is deployed and tested successfully under safe and realistic conditions with entirely new data. | The data for the field test was not used during the model development phase (e.g., new data acquisition after model development).<br><br>Performance requirements as specified by domain experts and auditors are fulfilled.<br><br>The field test is thoroughly documented.<br><br>CL 3 and higher: An independent third party (auditor, customer, etc.) must validate and supervise the field test demonstration. |
| 1 | Feature engineering | If applicable: Feature engineering methods are adequate and feasible. | The feature engineering is adequate and feasible for the given data and task (checked by domain and ML experts).<br>The feature engineering is thoroughly documented (e.g., motivation, methods, implementation). |
| 2 | ML implementation | Methods are implemented correctly, and the implementation fulfills basic quality requirements. | All algorithms are implemented correctly according to literature.<br>A state-of-the-art ML framework such as PyTorch, TensorFlow, Scikit-learn, or Keras is used. Alternatively, it is justified why the ML algorithms are implemented in-house.<br>The implementation fulfills basic requirements of modern software development (e.g., code structure, quality of implementation, documentation, version control).<br>Potential security issues, bugs, and other issues regarding the implementation are resolved. |

## 4.7 Machine Learning Scenarios with Pitfalls

As we now presented the most important aspects of our audit catalog for ML applications, we demonstrate how it may be applied in practice. Therefore, we discuss some possible pitfall-scenarios and how these issues can be captured by applying the audit catalog.

### 4.7.1 Misleading Performance Requirements

Choosing misleading performance requirements can invalidate the success of an ML project. For example, if a binary classifier is validated with the accuracy metric and the distribution of classes is imbalanced, this scenario can lead to a deceptively optimistic result. As an example of this effect we consider the detection of an illness. Usually the positive rate is quite small, thus let us assume that 100 out of 10000 patients test positive for the illness. If our trained model always predicted that a patient is healthy, the model would have an accuracy of 99%. So even though 99% might sound strong, such a model would obviously be completely useless. Overall accuracy is a misleading performance measure in this case, recall or miss-rate would be more appropriate. The audit catalog contains a point for the auditors (ML expert and/or domain expert) to check whether the defined minimum performance requirements under the given task are meaningful and reasonable.

## 4.7.2 Incorrect Data Splits via Wrong Oversampling

A common mistake is the incorrect handling of data during preprocessing. The correct workflow, specifically the order of operations on the dataset, is essential to avoid serious negative consequences.

In their work, Vandewiele et al. [93] analyze multiple papers, which report a near-perfect performance on a public dataset called *Term/Preterm Electrohysterogram database*. The data includes recordings of female patients and whether they delivered their child term or preterm. The authors argue that all papers that deal with this dataset make the same methodological mistake of applying oversampling before splitting the data into training, validation, and test splits. A severe problem arises due to oversampling of underrepresented data points (preterm deliveries). Near-perfect performance is then achieved due to duplicate data points being present in the data splits. More formally, the problem can be formulated by $D_{test} \cap D_{train} \neq \emptyset$, which violates well-established ML theory. This typically leads to overfitting and decreased generalization of the learned models. Similar issues may occur when information from the test split leaks into hyperparameters for data augmentation or if the developers analyze the test set during model development.

The problem described above can be detected when applying our audit catalog. For example, auditors (ML expert and/or domain expert) check whether the data and the splits are meaningful in terms of i.i.d. assumptions and whether the corresponding data processing algorithms are implemented correctly – which includes inspections regarding data leakage. Furthermore, for a CL of 2 or higher, we ensure that a trained model is tested under realistic conditions before being deployed. Note that since real-world datasets are often huge and wrong data splits cannot be detected easily, this aspect of the audit is performed in a qualitative manner currently.

## 4.7.3 Data Leakage

A different type of data leakage can occur if model inputs $x'$ include their corresponding labels ($x \in X, z \in Z, x' \in X \times Z$), which may lead to learning the identity function of the label $z$ and high test performance. Simply said, a weather prediction model may learn that "it rains on a rainy day". This scenario may render the model unusable for inference since labels are per definition most often unknown for new data. However, the issue may not always be as obvious as it seems. A real-world example of this scenario can be found in a certain Kaggle challenge about classification of marine animals via audio signals[23]. In this challenge, a team was able to achieve almost 100% accuracy by merely training the model based on the audio file lengths, time stamps and the chronological ordering of files.

Unfortunately, this kind of data leakage is in many cases not easy to detect. In the audit catalog we specifically ask for this issue to be examined. Moreover, input and output features have to be defined precisely. Furthermore, obvious correlations between the model input and output may be detected via interpretability methods.

## 4.7.4 Model Class Selection

Another mistake commonly encountered during model development is the incorrect choice of the model class. In ML there are typically many ways to solve a given task, leaving developers with many design choices to be made regarding the model class. A too low model class capacity becomes apparent during development due to unreachable performance requirements and can often be resolved easily. On the other hand, overly complex model classes (e.g., using over-parameterized ANNs) can lead to overfitting, which is often more difficult to identify in contrast to underfitting. Overfitting is characterized by poor test set performance in comparison to the training set performance (see Figure 11). This discrepancy can have different reasons but is most often caused by learning to exploit or memorize specific properties of individual data points in the training set.

---

[23] The ICML 2013 Whale Challenge – Right Whale Redux: https://www.kaggle.com/c/the-icml-2013-whale-challenge-right-whale-redux/discussion/4865#25839

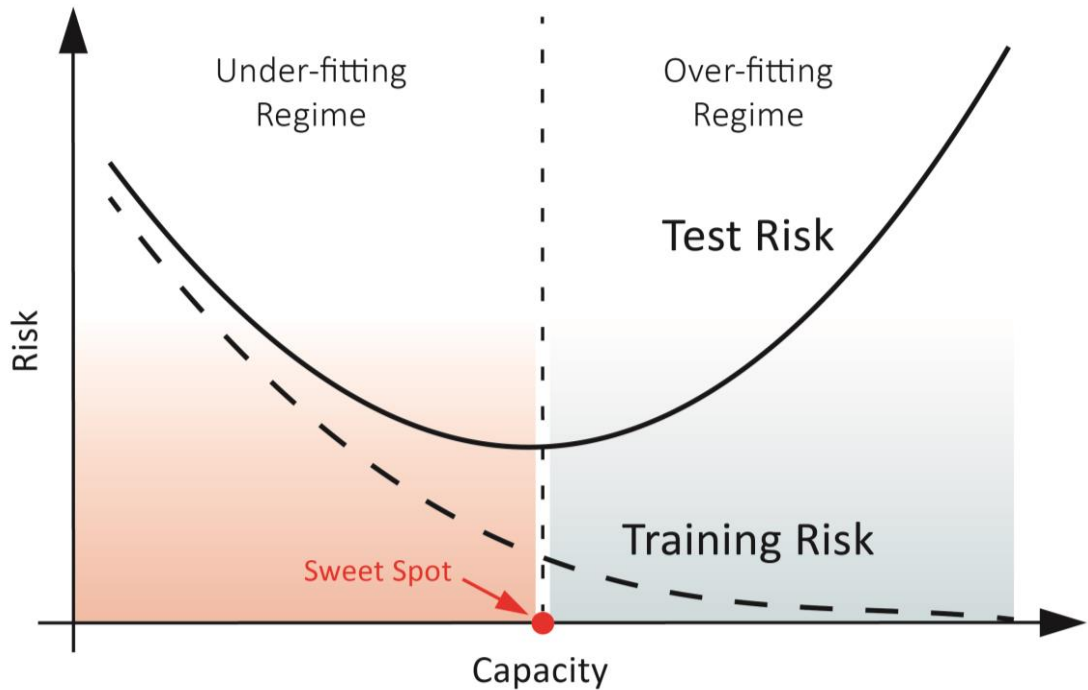

Figure 11: Classical bias-variance trade-off. Typically, there is a sweet spot for the model capacity which minimizes the test risk after training. Sub-optimal model capacities often lead to underfitting or overfitting.

In the classical ML context, the bias-variance trade-off [94] plays an essential role for choosing a model class with adequate capacity. Note that recent studies reveal more complicated underlying processes regarding the bias-variance tradeoff for large DL models and should be considered for in-depth model assessment [95, 96].

Moreover, there are other aspects regarding the model class which are worth to consider:

- Domain knowledge may be incorporated into the model class via algorithmic biases (e.g., convolutional neural networks for images or recurrent neural networks s for sequences). These algorithmic biases should be well-motivated and possible ablation studies should be performed to justify the use of individual model components.
- Model classes that are easier to interpret (such as decision trees) often receive more public acceptance compared to, e.g., ANNs.
- High-capacity models may be more costly to maintain [97] and may require significant energy consumption [98].

The audit catalog contains entries to investigate the model class selection and the adequate model capacity considering the given task.

## 4.7.5 Wrong Loss Functions

The choice of both an appropriate loss function and an appropriate evaluation metric is essential. For example, multi-class classification tasks require the use of the cross-entropy loss rather than, e.g., the MSE loss. Although the model might learn well using the MSE loss, the underlying mathematical assumptions are wrong, which can lead to unwanted effects and sub-optimal performance. The cross-entropy loss requires model outputs that can be interpreted as probability distributions (e.g., a softmax-activated vector) in order to be mathematically consistent. Complex tasks, e.g., for an automated driving system, might require several loss terms (e.g., regarding consistency) to achieve the desired model behavior. The audit catalog contains entries to check the used loss functions and evaluation metrics.

# 5 Conclusion and Outlook

The field of ML is rapidly evolving, with new ML applications increasing to influence science, industry, and our daily lives. As safety-critical ML applications begin to emerge, there is a need for creating legal liabilities in the form of norms, standards, and certificates. These tools allow for building reliance and lead to broad acceptance of ML systems in the public.

In this work, we motivate the need for a certification procedure by considering ongoing standardization activities, discussing issues regarding reproducibility, and identifying major challenges which have to be tackled. We propose an audit catalog for ML applications and lay the foundation for certification of low-risk ML applications within the setting of supervised learning. The catalog is designed to evaluate and verify ML applications from different viewpoints, such as secure software development, functional requirements, ethics, and data protection. We map the criticality of an ML application regarding the effects of its decisions, on people, environment, and companies via criticality levels and consider these aspects in the audit catalog. We are convinced that our audit catalog is the first step towards reliable certification of ML applications on the market and serves as a solid basis for future comprehensive certification of AI applications.

An audit catalog is an actively managed document and is subject to constant modification, which is guided by market demands and field experience. Existing points of the catalog may be modified, refined, or formalized to include more quantitative rather than qualitative checks. In the future, the catalog may be extended to include and handle a wider variety of ML approaches. Another possible direction of improvement is aiming towards certification of high-risk ML applications, such as automated driving and robotic agents in workspaces shared with humans. Studying topics such as data integrity, stability, adversarial attacks, and interpretability methods, may play an essential role in these further developments of the certification process.

# 6 Partners

## 6.1 Project Partners

**TÜV AUSTRIA Group**

TÜV AUSTRIA is an international company with branches in more than 20 countries of the world. TÜV AUSTRIA employs about 2.000 employees. The service competencies of the four business areas „Industry & Energy", „Infrastructure & Transportation", „Business Assurance", and „Digital Services" encompass the areas of testing, monitoring, certification, education, training and consulting. For almost 150 years now, we have been accompanying and securing technical innovations. Thus, our mission is to improve safety, security, and quality in a sustainable way.

From its offices in Cologne and Vienna, TÜV AUSTRIA Group Member TÜV TRUST IT is the neutral, objective and independent partner for the industry with regard to information security and data privacy. The mission is to support companies in protecting their information assets. Information assets comprise all relevant data within a company, which is necessary to ensure that business operations run properly and are provided via IT infrastructure and processes. Information values are thus assets that, like all other corporate assets, need to be protected in accordance with their importance. The services of TÜV TRUST IT are based on internationally recognized standards and best practices.

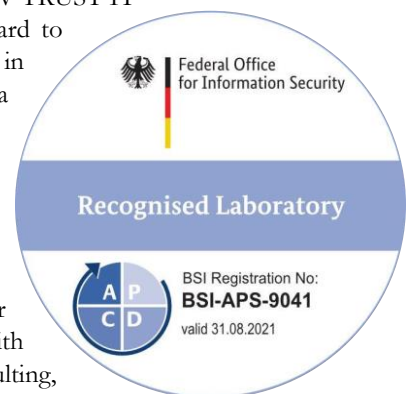

In addition, TÜV TRUST IT is certified by the German Federal Office for Information Security (BSI) as an IT security service provider in accordance with article 9, section 2 BSIG for the scope of information security revision, consulting, and penetration testing. This means that it has the necessary competencies and meets the relevant criteria of DIN EN ISO/IEC 17025:2005.

Contact: digitalservices@tuv.at

**Johannes Kepler Universität Linz (JKU) − The Institute for Machine Learning (IML)**

At just 50 years of age, the Johannes Kepler University Linz is a comparably young university. Considering its young age, the university's accomplishments over the past five decades have been all the more remarkable. Today, the JKU Linz is home to approximately 3,300 employees and 21,000 students. Their curiosity, creativity, and ingenuity are a testament to the university's namesake, Johannes Kepler, who lived and worked in Linz between 1612 and 1626. The JKU Linz brings the past and the present together and, as Upper Austria's largest institution for research and teaching, the university is also paving the way to the future.

The Institute for Machine Learning (IML) at the JKU conducts internationally renowned research and offers a sound education in Machine Learning with the newly established AI study program. The IML's latest research deals, for example, with the development of Machine Learning algorithms in the field of NLP, Few-Shot-Learning, Reinforcement-Learning and with the translation to Life Sciences and Healthcare. As of 2021, the IML and the LIT AI Lab have about 50 employees in total.

The LIT AI Laboratory (LIT AI Lab) headed by Prof. Sepp Hochreiter was founded as a permanent research center of the Linz Institute of Technology (LIT). In the unique environment offered by the Johannes Kepler University (JKU) Linz, the LIT AI Lab bundles JKU's world-class expertise in artificial intelligence (AI) for shaping and advancing AI research and its industrial applications. The LIT Lab is committed to scientific excellence. Its focus is on theoretical and experimental research in Machine Learning, logical reasoning, and computational perception. The next generations of AI researchers and engineers are educated at various academic levels in AI technology.

The LIT AI Lab was chosen as the ELLIS Unit Linz of the European Laboratory for Learning and Intelligent Systems (ELLIS). ELLIS stands for excellence in Machine Learning and is a European network, which promotes excellent research and also aims to boost economic growth in Europe by leveraging AI technologies. The ELLIS unit Linz will contribute to coordinating Machine Learning excellence in Europe and to establish a local sustainable ecosystem of Machine Learning stakeholders covering the entire value network to facilitate and accelerate a broad uptake and integration of Machine Learning technologies. The unit will conduct basic Machine Learning research at the highest levels in coordination with other ELLIS sites and thereby advance theories, algorithms, and applications of Machine Learning. The unit will be established on the premises of the LIT AI Lab located at the Johannes Kepler University Linz (JKU).

Contact: secretary@ml.jku.at

## 6.2 Acknowledgements

We kindly thank all reviewers for participating in the soft-review process. We also thank Andreas Mayr, Sebastian Lehner, Lukas Gruber, Markus Holzleitner, Günter Klambauer, Doris Kaiserreiner, Johannes Lehner, and Vihang Patil for their valuable feedback to improve the quality of this work.

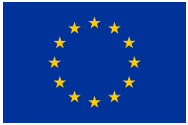

"This project has received funding from the European Union's Horizon 2020 research and innovation programme under grant agreement No 951847".

# 8 Appendix

## 8.1 History of AI

In the late 1930s, 1940s, and early 1950s, scientists from various fields, such as mathematics, psychology, and engineering, began discussing thinking machines, or the creation of an artificial brain. The reason was that new research in neurology showed that the brain is an electrical network of neurons firing in all-or-none pulses [99]. For example, Norbert Wiener's cybernetics, Claude Shannon's information theory, and Alan Turing's computational theory suggested that it might be possible to construct an electronic machine that simulates the cognitive capabilities of the human brain [100]. In 1943, Walter Pitts and Warren McCulloch presented their theory that employed logic and the mathematical notion of computation to explain how neural mechanisms might realize mental functions [101]. Pitts and McCulloch were the first to describe what later researchers would call a neural network [102].

From 1950 to 1960, researchers focused on the development of games and what is known as "pattern recognition". This is the process of analyzing images, speech segments, electronic signals, or other data samples and then classifying them into one of several categories. The actual beginning of AI can be dated back to the workshop "Dartmouth Summer Project on Artificial Intelligence" in the summer of 1956. From the mid-1960s to the mid-1970s, AI research flourished, and several new research groups formed at universities and in companies. During this time, AI experienced a strong upswing, which led to many new discoveries and to the advent of research fields like computer vision (e.g., face recognition), knowledge representation and reasoning (e.g., logic programming languages like Prolog), and NLP [103, 104].

In the second half of the 1970s, AI research was heavily criticized because AI failed to solve real-world problems. The high expectations of AI could not be met and, as a result, most funding discontinued. This period is called the first AI winter [105]. However, AI methods and computer hardware became more powerful and AI researchers were able to shift their focus to specific problems and application domains. During the first AI winter researcher specialized in sub-disciplines such as natural language processing and speech recognition, expert systems (e.g., the consulting system Mycin[24]), and computer vision (e.g., image understanding).

In particular, expert systems were built up by the means of human knowledge using conditional computer programming, resulting in large knowledge bases. These knowledge bases allowed for logical reasoning and effective searching, achieving some notable results and raising great expectations about the possibilities of AI. However, the effort involved in capturing human knowledge, representing it via knowledge bases, and also the maintenance of such knowledge bases turned out to be immense and impracticable. Since overall the difficulties outweighed the successes of AI and it generally failed to live up to expectations and promises, the period from around the mid-1980s to the end of the 1980s marked the second AI winter [106].

ML research emerged as the largest subfield of AI and was able to revive AI again after the second winter in the early 1990s. The rise of ML was initiated by deepening its underlying engineering science and mathematics. As a result, sophisticated new engineering tools emerged and others were refined, significantly increasing the overall power of ML systems and, thereby, invalidating some of the earlier criticisms. In recent decades, ML has become much more advanced and now includes an impressive suite of powerful computational tools that help to solve many real-world problems. The reason why these tools can be used with such effectiveness, is the increasing computing capacity of relatively inexpensive computers, the availability of large datasets, and the recent developments in DL [22]. Today's AI programs closely approximate human cognitive abilities for many tasks and even outperform humans in some tasks. The relationship between AI, ML, and DL is visualized in Figure 3 in the main text.

## 8.2 Selected Standardization Activities

Table 3 presents selected examples of worldwide standardization activities in the areas of reliability and robustness, safety, ethics, fairness and non-discrimination, and human Agency and oversight. It should be noted that the listing is done without judgment or evaluation and that the different documents may not reflect our opinion.

---

[24] Mycin. In Wikipedia, The Free Encyclopedia, https://en.wikipedia.org/wiki/Mycin

Table 3: Overview of selected AI standardization activities worldwide [107].

| Worldwide AI Standardization Activities by Categories | | |
|---|---|---|
| **Document** | **Title** | **Short Description** |
| **Reliability & Robustness** | | |
| ISO/IEC NP 24029-1 and 24029-2 | AI – Assessment of the robustness of neural networks | Part 1 (24029-1): Overview<br>Part 2 (24029-2): Formal methods methodology |
| ITU-T F.AI-DLFE | Deep Learning Software Framework Evaluation Methodology | Requirements for architectures of Deep Learning |
| ITU-T F.AI-DLPB | Metrics and evaluation methods for deep neural network processor benchmark | Evaluation scheme for Deep Learning with regards to interference, training, application, network, and processor |
| ETSI DTR INT 008 (TR 103 821) | Autonomic network engineering for the self-managing Future Internet (AFI); AI in test Systems and Testing AI models. | AI in test systems, testing AI models and the ETSI GANA model's cognitive decision elements via a generic test framework for testing ETSI GANA multi-layer autonomics and their AI algorithms for closed-loop network automation |
| DIN SPEC 92001-1 and 92001-2 | AI – Life Cycle Processes and Quality Requirements | Part 1 (92001-1): Quality Meta Model<br>Part 2 (92001-2): Robustness |
| **Safety** | | |
| ISO/CD TR 22100-5 | Safety of machinery – Relationship with ISO 12100 – Part 5: Implications of embedded AI-ML | The technical report describes how hazards associated with the use of ML systems in machines should be considered in the risk management process. |
| ISO 26262:2018 | Road vehicles – Functional safety | Provides an automotive safety lifecycle, i.e., management, development, production, operation, service, decommissioning, and safety-oriented analysis (ASIL). |
| IEEE P2802 | Standard for the Performance and Safety Evaluation of AI Based Medical Device: Terminology | The standard establishes the terminology used in AI medical devices, including definitions of fundamental concepts and methodology, safety, efficacy, risks, and quality management. |
| ISO/IEC AWI TR 5469 | AI – Functional safety and AI systems | |
| **Ethics** | | |
| ISO/IEC AWI TR 24368 | Information technology – AI – Overview of ethical and societal concerns | Technical report on ethical and societal challenges of using AI. |
| IEEE P70xx series | "Standards for the future of ethically aligned autonomous and intelligent systems" | The IEEE P70xx series aims to translate the principles of the IEEE Ethically Aligned Design: A Vision for Prioritizing Human Well-being with Autonomous and Intelligent Systems (version 2, 2017) document into actionable guidelines or frameworks that can be used as practical industry standards. |
| **Fairness & Non-Discrimination** | | |
| ISO/IEC AWI TR 24027 | Information technology – AI – Bias in AI systems and AI aided decision making | Technical report to describe bias in AI-systems |
| **Human Agency & Oversight** | | |
| IEEE 7010-2020 | IEEE Recommended Practice for Assessing the Impact of Autonomous and Intelligent Systems on Human Well-Being | The impact of AI or autonomous and intelligent systems (A/IS) on humans is measured by this standard. |

## 8.3 Common Evaluation Metrics in ML

Metrics play an essential role in supervised learning. Thus, we provide a brief overview of commonly used metrics in literature.

### 8.3.1 Metrics for Classification-related Tasks

The confusion matrix compares (the number of) predicted classes versus actual classes for all k classes. The matrix allows for an in-depth study of misclassifications; true positives (TP), false positives (FP), true negatives (TN), and false negatives (FN). The confusion matrix allows for calculating sensitivity (also called recall, hit rate, or true positive rate, TP/P), specificity (also called selectivity or true negative rate, TN/N), precision (also called positive predictive value, TP/(TP+FP)), etc. It also allows for calculating common scores such as accuracy (TP+TN / (P+N)) or the F1 score (the harmonic mean of precision and sensitivity, 2TP/(2TP+FP+FN)). Cohen's Kappa is an adjusted form of the accuracy that considers the accuracy that would have happened by random guesses. The accuracy ((TP+TN)/(FP+FN+TP+TN)) is one of the most important evaluation measures. The *error rate* is defined as 1-ACC.

The receiver operating characteristics (ROC) curve describes the balance between sensitivity and specificity of a model with different prediction probability cutoffs (different classification thresholds of the sigmoid unit) for binary classification tasks (logistic regression). We would like to have a model that has both high sensitivity and specificity, but in practice we often have to accept a certain tradeoff, or we want to set some task-specific importance of one component over the other. (e.g., we would like to correctly classify all persons with a certain disease (TP) while accepting the possibility to get more FP; therefore, we lower the threshold.) The ROC curve is the only metric that measures how well the model does for different cutoff values. The Area under the ROC curve (AUC) can be used as a single-valued goodness score and allows to compare different ROC curves. Note that for imbalanced data (e.g., rare diseases) one may prefer using precision instead of the false positive rate (1 - specificity) to generate ROC curves. Moreover, ROC and AUC is used to evaluate multiclass classification problems with imbalanced data ( [108] ).

For multiclass (softmax output) and multilabel (sigmoid output) classification tasks, the notions of precision, recall, and F-measures can be applied to each label independently. Typically, average statistics are reported.

### 8.3.2 Metrics for Regression-related Tasks

In general, errors are measured by the means of the distance between the model output and the target output. Typical metrics include the mean absolute error (MAE or L1 distance), the mean squared error (MSE or L2 distance), or the root mean squared error (RMSE). Note that MSE penalizes outliers stronger compared to MAE. If RMSE is close to MAE, the model makes many relatively small errors. If RMSE is close to MSE, the model makes few but large errors. Note that MSE assumes a Gaussian noise distribution, whereas MAE assumes a Laplace distribution. The max error captures the worst-case error a model can make. The explained variance score describes how well the variance of the data can be modelled. The $R^2$ score indicates the goodness of a fit and therefore a measure of how well unseen samples are likely to be predicted by the model. For neural network training, MSE may be a good choice by default. However, MAE (or even binary cross-entropy) can yield better results for certain tasks ( [109] ).

### 8.3.3 Domain-specific Evaluation Metrics

Different ML-domains often use specialized evaluation metrics. In the following, we briefly list some of the most prominent metrics within these domains.

**Vision:** Detection: Average precision. Classification: Top-1 score (target label is highest ranked predicted class) or Top-5 score (target label is one of the top 5 highest ranked predicted classes) ( [26] ). Segmentation: Intersection over Union (IoU, aka. Jaccard index) ( [110] ) or Dice Coefficient.

**Life-Science:** Classification: Balanced accuracy, ROC, AUC, F1 score, precision, recall, Cohen's Kappa. Drug discovery: validity, novelty, uniqueness, solubility, hit rate.

**Natural Language Processing:** perplexity, BLEU score ( [111] ), accuracy.

**Generative models:** negative log-likelihood, Frechet Inception Distance (FID, [112]), Inception Score (IS, [113]), Frechet Video Distance (FVD, [114]). Frechet ChemNet Distance (FCD, [115]), Fréchet Audio Distance (FAD, [116]), Kullback–Leibler divergence (KL), Wasserstein distance.